\documentclass[lettersize,journal]{IEEEtran}

\usepackage{algorithm}
\usepackage{algorithmic}
\usepackage{amsfonts}
\usepackage{amsmath}
\usepackage{amssymb}
\usepackage{array}
\usepackage{booktabs}
\usepackage{cite}
\usepackage{enumitem}
\usepackage{fontawesome5}  % 更现代，支持 \faRuler、\faSearch 等
\usepackage{graphicx}
\usepackage{hyperref}
\usepackage{longtable}
\usepackage{makecell}
\usepackage{multirow}
\usepackage{multicol}
\usepackage{subcaption}
\usepackage[most]{tcolorbox}
\usepackage{tikz}
\usepackage[table]{xcolor} % 引入 xcolor 包，并启用表格支持
\usepackage{colortbl}

\usepackage{mdframed}
\usepackage{caption}
\newcommand{\hhline}{%
    \noalign {\ifnum 0=`}\fi \hrule height 1pt
    \futurelet \reserved@a \@xhline
}

\definecolor{lightgray2}{gray}{0.85}

\definecolor{myred}{RGB}{237,28,80 } % 更鲜艳的红色
\definecolor{scarlet}{RGB}{255,36,0} % 猩红
\definecolor{keywordred}{RGB}{200,34,34}
\definecolor{iceblue}{RGB}{214, 230, 245}    % 冰蓝
\definecolor{pastelyellow}{RGB}{254, 240, 158} % 浅鹅黄
\definecolor{creamyellow}{RGB}{255,246,213}

\usepackage{xspace}
\newcommand{\ourmethod}{DSIPA\xspace}
\newcommand{\ourmethodTT}{\texttt{\textit{\textbf{\ourmethod}}}\xspace}

\newcommand{\GPTThree}{\textit{Text-Davinci-003}\xspace}
\newcommand{\GPTThreeFiveTurbo}{\textit{GPT-3.5-turbo}\xspace}
\newcommand{\GeminiOneFivePro}{\textit{Gemini-1.5-pro}\xspace}
\newcommand{\GPTFourAll}{\textit{GPT-4-0613}\xspace}
\newcommand{\GPTFiveTwoAll}{\textit{GPT-5.2}\xspace}
\newcommand{\LLaMa}{\textit{LLaMa-3.3}\xspace}
\newcommand{\Claude}{\textit{Claude-3}\xspace}

\newcommand{\datasetnews}{\textit{Reuter News}\xspace}
\newcommand{\datasetcode}{\textit{HumanEval Code}\xspace}
\newcommand{\datasetessay}{\textit{Student Essay}\xspace}
\newcommand{\datasetpaper}{\textit{Academic Paper}\xspace}
\newcommand{\datasetreview}{\textit{Yelp Review}\xspace}

\begin{document}

\title{\ourmethod: Detecting LLM-Generated Texts via Sentiment-Invariant Patterns Divergence Analysis}

\author{
\large Siyuan Li, Aodu Wulianghai, Guangyan Li, Xi Lin, \IEEEmembership{\large Member, IEEE}, Qinghua Mao, Yuliang Chen, \\ Jun Wu, \IEEEmembership{\large Senior Member, IEEE}, and Jianhua Li, \IEEEmembership{\large Senior Member, IEEE}
    \thanks{Siyuan Li, Xi Lin, Qinghua Mao, Yuliang Chen, Jun Wu, and Jianhua Li are with the School of Computer Science, Shanghai Jiao Tong University, Shanghai, China, and also with Shanghai Key Laboratory of Integrated Administration Technologies for Information Security, Shanghai, China 
    (Email: \{siyuanli, linxi234, mmmm2018, chenyuliang, junwuhn, lijh888\}@sjtu.edu.cn).}
    \thanks{Aodu Wulianghai is with the School of Computer Science, Shanghai Jiao Tong University, Shanghai, China (Email: melusine.wlhad@sjtu.edu.cn).}
    \thanks{Guangyan Li is with the State Key Laboratory of Multimodal Artificial Intelligence Systems, Institute of Automation, Chinese Academy of Sciences, Beijing 100190, China (Email: liguangyan2022@ia.ac.cn).}
}

\markboth{Submitted to IEEE Transactions on Dependable and Secure Computing}
{Shell \MakeLowercase{\textit{et al.}}: A Sample Article Using IEEEtran.cls for IEEE Journals}

\IEEEpubid{0000--0000/00\$00.00~\copyright~2026 IEEE}

\maketitle
\begin{abstract}
The rapid advancement of large language models (LLMs) presents new security challenges, particularly in detecting machine-generated text used for misinformation, impersonation, and content forgery. 
Most existing detection approaches struggle with robustness against adversarial perturbation, paraphrasing attacks, and domain shifts, often requiring restrictive access to model parameters or large labeled datasets. 
To address this, we propose \ourmethodTT, a novel training-free framework that detects LLM-generated content by quantifying sentiment distributional stability under controlled stylistic variation.
It is based on the observation that LLMs typically exhibit more emotionally consistent outputs, while human-written texts display greater affective variation. 
Our framework operates in a zero-shot, black-box manner, leveraging two unsupervised metrics, \textit{sentiment distribution consistency} and \textit{sentiment distribution preservation}, to capture these intrinsic behavioral asymmetries without the need for parameter updates or probability access.
Extensive experiments are conducted on state-of-the-art proprietary and open-source models, including \GPTFiveTwoAll, \GeminiOneFivePro, \Claude, and \LLaMa. 
Evaluations on five domains, such as news articles, programming code, student essays, academic papers, and community comments, demonstrate that \ourmethod improves F1 detection scores by up to 49.89\% over baseline methods.
The framework exhibits superior generalizability across domains and strong resilience to adversarial conditions, providing a robust and interpretable behavioral signal for secure content identification in the evolving LLM landscape. 
\end{abstract}


\section{Introduction}
\IEEEPARstart{T}{he} proliferation of large language models (LLMs) has transformed content creation, enabling the automatic generation of fluent, coherent, and contextually relevant text across diverse domains~\cite{lei2025pald,jiang2024survey,LLMSurvey1:yang2023harnessing}.
These capabilities have rapidly accelerated the deployment of LLMs in education, journalism, customer service, and medical assistance, offering clear benefits in productivity, personalization, and communication~\cite{zhou2024survey,li2024trustworthy,chu2025llm,liu2025qos,su2025large}.
However, the development also poses growing threats to information integrity and digital security.
In contexts such as misinformation propagation, academic fraud, identity spoofing, and large-scale social engineering, machine-generated text can be misused to impersonate humans or fabricate plausible and deceptive narratives~\cite{lee2023language,sp2023:pu2023deepfake,li2026honeytrap,wu2024detectrl}.
As LLMs evolve toward human-level stylistic fluency and coherence, the boundary between authentic and synthetic content becomes increasingly blurred~\cite{PNAS:jakesch2023human, cornelius2024bust}, creating significant risks for secure communication, online trust, and forensic auditing.
This creates an urgent demand for reliable and generalizable detection mechanisms that can operate in high-stakes security environments without relying on model access or assumptions about generation provenance~\cite{chen2025online}.
\IEEEpubidadjcol
% \textcolor{keywordred}{
\begin{figure}[!t]
    \centering
    \includegraphics[width=\linewidth]{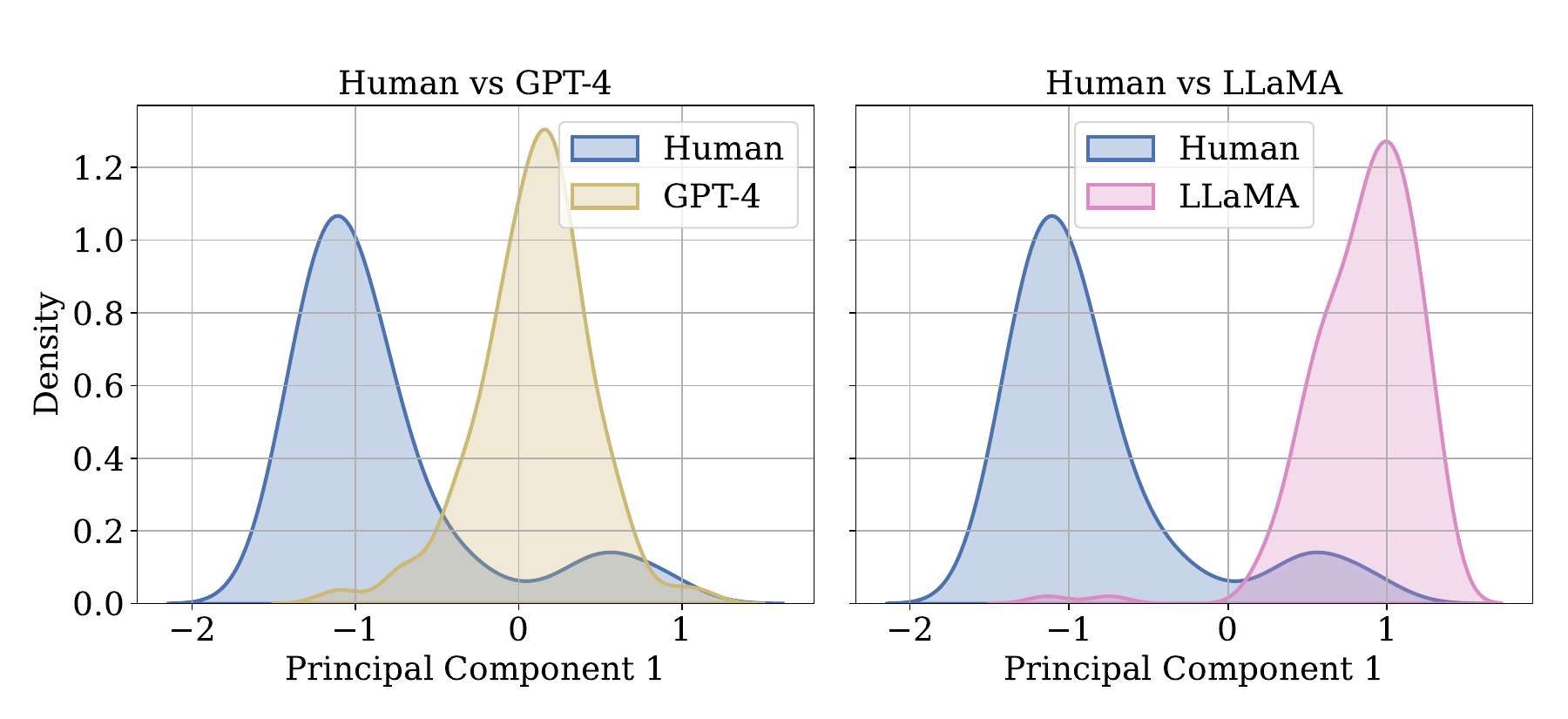}
    \caption{A schematic comparison of the \textcolor{black}{\textit{Sentiment Distribution Consistency}} between human-written and LLM-generated texts within the review domain.     Each subplot illustrates the density projection along the first principal component (PC1), with each point representing an embedded paragraph of no more than 64 tokens. 
    \textit{The sentiment distribution feature clearly separates LLM-generated content from human-written ones.}
    }
    \label{figure:scatter-plot}
\end{figure}

To address this challenge, recent research has proposed a variety of methods for distinguishing LLM-generated content from human-written text~\cite{DetectGPT:mitchell2023detectgpt, RADAR:hu2024radar, canAI-generated?:sadasivan2023can}.
These approaches generally fall into three broad categories: watermark-based detection~\cite{watermark0:abdelnabi2021adversarial, watermark2:fu2023watermarking, Watermark3-kirchenbauer2023reliability}, supervised classification-based detection~\cite{OpenAI-CLS2019, StyleRepresentations:soto2024few}, and statistical feature-based analysis~\cite{li2025styledecipher, bao2024fast, tian2024Multiscale}.
While each of these paradigms shows promise under constrained conditions, they face serious limitations in adversarial or black-box scenarios~\cite{RADAR:hu2024radar}.
Watermark signals can be removed or become ineffective after paraphrasing, and are entirely absent in generations from open-source LLMs~\cite{Retrieval:krishna2024paraphrasing}; classification-based methods often overfit to specific model outputs and struggle to generalize across domains or stylistic variation; statistical methods frequently require token probability access, which is unavailable for many real-world deployments~\cite{DetectGPT:mitchell2023detectgpt, bao2024fast, li2026model}.
More fundamentally, most existing methods treat detection as a surface-level classification task, overlooking deeper behavioral discrepancies between human and machine-generated content~\cite{RADAR:hu2024radar, DNA-GPT:yang2023dna}.
This absence of behavior-grounded modeling limits interpretability, adaptability, and long-term robustness, which are the key requirements for practical deployment in secure content attribution systems.

Unlike previous approaches, we explore a distinct viewpoint: \textit{can the emotional patterns embedded within language be leveraged to determine authorship via a training-free mechanism, without the need for supervised learning, parameter updates, or access to proprietary model parameters?} Inspired by this inquiry, we perform a series of empirical analyses and find that the consistency of emotional expression in text acts as a powerful and unique indicator for authorship attribution, which is captured by the constancy and fluctuation of sentiment. 
Our key finding is that LLM-generated content often exhibits relatively stable sentiment distributions, irrespective of genre or stylistic shifts. In this paper, we hypothesize that this behavior arises from the nature of LLM training, where exposure to diverse and carefully curated datasets, combined with optimization strategies focused on minimizing risk, leads to outputs that are emotionally neutral and stylistically consistent~\cite{SAforLLMWWW:deng2023llms, naveed2023comprehensive, vijay2025neutral, shen2024donowcharacterizingevaluating}. 
In contrast, human-written texts are naturally infused with more diverse and fluctuating emotional dynamics, even among texts that are similar in topic or function~\cite{SAsurvey1:zhang2022survey, LLMSurvey2:hu2023survey}. These subtle yet measurable differences in sentiment variability provide an underexplored signature for detecting LLM-generated texts.

 Based on this insight, we propose \ourmethodTT, a training-free framework that detects machine authorship by measuring sentiment-invariant divergence under controlled rewriting. Unlike classifier fine-tuning, token-level likelihoods, or watermark cues, \ourmethod relies only on distributional shifts in sentiment expression and supports robust zero-shot deployment across proprietary and open-source LLMs. We evaluate \ourmethod on texts generated by multiple LLM families, including \GPTThreeFiveTurbo, \GPTFiveTwoAll, \GeminiOneFivePro, and \Claude. 
Our experiments show that sentiment stability provides a generalizable, interpretable, and resilient behavioral signal for secure authorship inference. 
Notably, the framework demonstrates strong zero-shot performance and sustains high accuracy under adversarial conditions such as paraphrasing, domain shifts, and input-length variations.

% Based on this insight, we propose the \ourmethodTT, a novel framework for analyzing sentiment-invariant patterns as a signal of machine authorship. 
% Unlike conventional detection systems that depend on classifier fine-tuning, token-level likelihoods, or watermark-based embeddings, \ourmethod instead operates by quantifying distributional variance in sentiment expression under controlled stylistic variation. 
% This approach enables robust detection across both proprietary and open-source models, and generalizes across diverse content types without reliance on model-specific heuristics.
% We perform comprehensive evaluations on texts generated by multiple LLM families, such as \GPTThreeFiveTurbo, \GPTFourAll, \GeminiOneFivePro, and \Claude, and across five representative domains including misinformation detection, academic writing, program code, student essays, and online reviews. 
% Our experiments show that sentiment stability provides a generalizable, interpretable, and resilient behavioral signal for secure authorship inference. 
% Notably, the framework demonstrates strong zero-shot performance and sustains high accuracy under adversarial conditions such as paraphrasing, domain shifts, and input-length variations.

Our main contributions can be summarized as follows:

\begin{itemize}
    \item \textit{\textbf{Sentiment stability under stylistic variation.}} 
    We introduce a new behavioral perspective for LLM-generated text detection by leveraging the stability of sentiment patterns across stylistic variants. This insight reframes sentiment invariance as a fundamental signal of machine authorship, offering a principled alternative to classifier-based or watermarking approaches.
    
    \item \textit{\textbf{Training-free sentiment divergence metrics.}} 
    We develop a model-agnostic and training-free framework that quantifies emotional consistency through two unsupervised metrics: \textit{sentiment distribution consistency} and \textit{sentiment distribution preservation}. The proposed mechanism eliminates the need for labeled data, model access, or backpropagation-based optimization, relying instead on intrinsic behavioral properties of LLMs.
    
    \item \textit{\textbf{Zero-shot and robust detection across models and domains.}} 
    We conduct extensive zero-shot evaluations across diverse LLM families, datasets, and content domains, demonstrating that DSIPA achieves robust performance through task-agnostic calibration of distributional boundaries. The results highlight the practical applicability of \ourmethod in security-sensitive scenarios, even under challenging adversarial conditions such as paraphrasing, domain shifts, and input-length variations.
\end{itemize}

% \begin{itemize}
%     \item \textit{\textbf{Sentiment stability under stylistic variation.}} We introduce a new behavioral perspective for LLM-generated text detection by leveraging the stability of sentiment patterns across stylistic variants, offering an alternative to classifier-based or watermarking methods.
%     \item \textit{\textbf{Sentiment divergence metrics without supervision.}} We develop a model-agnostic framework that quantifies emotional consistency and divergence through two unsupervised metrics, sentiment distribution consistency and sentiment distribution preservation, without requiring labeled training data or access to the models.
%     \item \textit{\textbf{Zero-shot and robust detection across models and datasets.}} We conduct extensive zero-shot evaluations across diverse models and domains, demonstrating robust performance even under challenging adversarial settings, and highlighting the practical applicability of \ourmethod in security-sensitive contexts.
% \end{itemize}

\section{Related Works}
The rapid advancement of large language models (LLMs) has significantly blurred the boundary between human-written and machine-generated text, raising urgent concerns in domains such as misinformation, authorship fraud, and secure content provenance~\cite{pearce2023examining, scherrer2024evaluating, AGIGPT-4:bubeck2023sparks}.
Before the emergence of powerful models like GPT-3~\cite{GPT-3} and LLaMa-2~\cite{LLaMA2}, earlier detection efforts focused on handcrafted features and statistical anomalies, such as entropy visualization in GLTR~\cite{GLTR:gehrmann2019gltr}, or shallow classifiers based on linguistic statistics~\cite{OpenAI-CLS2019, tay2020reverse, uchendu2020authorship}.
However, these traditional approaches are increasingly ineffective as LLMs generate more fluent, semantically consistent outputs~\cite{DetectionSurvey1:yang2023survey, wu2023survey}.

\subsection{Detection of LLM-Generated Texts}
Recent approaches for generated text detection fall into three main paradigms: watermarking, supervised classification, and statistical analysis~\cite{chakraborty2024position}.

\subsubsection{Watermarking Detectors}
Watermark-based methods aim to insert identifiable patterns or signals into the generated text during the decoding process, enabling posterior attribution.
Earlier efforts like adversarial watermarking transformers encode discrete binary sequences within generated content~\cite{watermark0:abdelnabi2021adversarial}, while recent techniques modify the sampling distributions to inject subtle statistical traces without significantly altering fluency or semantics~\cite{Watermark1-forLLM:kirchenbauer2023watermark}.
Such methods provide practical advantages under controlled deployment scenarios, where the generator is known and instrumentation is feasible.
However, despite their growing sophistication, watermarking strategies face critical limitations in open settings.
First, they are fragile under common text manipulations such as paraphrasing, compression, or semantic editing~\cite{Retrieval:krishna2024paraphrasing}, which often invalidate the watermark signal.
Second, watermarks are ineffective for legacy outputs or open-source model generations where watermark embedding is not available~\cite{Watermark3-kirchenbauer2023reliability}.
Finally, their dependence on access to the text-generation pipeline limits their applicability to forensic or zero-access use cases.

\subsubsection{Supervised Classification}
Adversarial learning has also been introduced to enhance robustness, exemplified by RADAR~\cite{RADAR:hu2024radar}, which fine-tunes detection models using synthetically perturbed data.
Despite their promise in controlled environments, classification-based methods face scalability and generalization issues.
They require large volumes of labeled training data that must be refreshed frequently as LLMs evolve.
In zero-shot settings or when applied to unseen models or domains, their performance often degrades sharply~\cite{StyleRepresentations:soto2024few}.
Moreover, they lack interpretability and typically offer limited insights into why a piece of text is flagged as machine-generated, which is a critical gap for high-stakes decisions such as content moderation or legal verification.

\subsubsection{Statistical Approaches}
Statistical detectors aim to identify intrinsic generation artifacts in LLM outputs.
One influential approach, DetectGPT~\cite{DetectGPT:mitchell2023detectgpt}, posits that LLM-generated texts exhibit distinct curvature in the log-likelihood landscape.
By comparing the likelihood curvature of a sample against its perturbed variants, DetectGPT identifies unstable regions that are characteristic of generated content.
Fast-DetectGPT~\cite{bao2024fast} builds on this method, offering a computationally optimized variant with reduced inference overhead.
 Similarly, Binoculars~\cite{baseline2:Binoculars:hans2024spotting} advances zero-shot detection by employing a dual-model scoring mechanism that contrasts observer and performer perplexities. 
It identifies machine-generated text with high accuracy without requiring sample perturbations.
R-Detect~\cite{baseline1:R-Detect:song2025deep} adopts a distribution-based approach that computes divergence measures over multiple candidate rewrites to detect subtle generation patterns.
Other methods explore alternative statistical signals.
Some leverage n-gram distributions or token entropy~\cite{zhang2024detecting}, while others propose metrics like intrinsic dimensionality~\cite{tulchinskii2024intrinsic} or maximum mean discrepancy to capture deviations from human writing norms.
Rewriting-based methods~\cite{baseline3:RAIDAR:mao2024detecting} generate paraphrased variants to examine response divergence as an indicator of generation origin.
While these techniques offer training-free benefits, many remain sensitive to surface-level text edits and depend on specific sampling conditions.
In addition, they often require multiple perturbed generations, which limits their efficiency in large-scale or real-time detection scenarios.

\subsection{Sentiment Analysis in the Era of LLMs}
Sentiment analysis has evolved from simple binary classification (i.e., distinguishing positive from negative opinions) into a multifaceted research field encompassing aspect-based sentiment detection, fine-grained opinion mining, and dynamic emotion trajectory analysis~\cite{SAsurvey1:zhang2022survey, SA1:dufraisse2023mad}.
Recent advancements in LLMs have drastically reshaped this landscape, with state-of-the-art models like GPT-4, Claude, and Gemini exhibiting strong zero-shot and few-shot capabilities in sentiment prediction tasks, even without extensive task-specific fine-tuning~\cite{zhang2024sentiment}. 
Several studies explore how LLMs compare with fine-tuned models.
Zhong \textit{et al.}\cite{zhong2023can} show that while LLMs rival traditional models in general sentiment tasks, they also demonstrate a marked tendency toward emotionally neutral outputs, especially when optimized under safety or alignment constraints\cite{li2023large, shen2024donowcharacterizingevaluating}.
Deng \textit{et al.}~\cite{SAforLLMWWW:deng2023llms} further leverage the powerful semantic understanding of LLMs for weakly supervised sentiment labeling, a critical task in scenarios where labeled data is scarce or costly to obtain.
Building on these findings, our work introduces a novel perspective: analyzing the sentiment stability across stylistically altered versions of the same text as a behavioral fingerprint for machine authorship. 
This method doesn't rely on extreme or categorical sentiment but instead measures emotional consistency across rewrites to reflect the affective rigidity typical of LLM outputs.

\subsection{Secure and Interpretable Provenance Attribution}
Recent efforts have highlighted the importance of provenance analysis that is both interpretable and robust to adversarial modifications.
Works such as DNA-GPT~\cite{DNA-GPT:yang2023dna}, Ghostbuster~\cite{verma2024ghostbuster}, and BUST~\cite{cornelius2024bust} adopt a behavior-based or representation-driven approach, leveraging stylistic, semantic, or activation-level features to infer authorship.
Recent studies in academic writing detection, such as Liu \textit{et al.}~\cite{10.1145/3711896.3737408}, have further emphasized the need for detectors that can generalize and adapt across diverse scholarly domains.
Complementary research in deepfake detection~\cite{sp2023:pu2023deepfake} explores resilience under paraphrasing, prompt injection, and adversarial transformations.
These methods share a common goal with ours: moving beyond opaque classifiers to detection grounded in interpretable and persistent behavioral signals.
However, most rely on embeddings or supervised classifiers, which may not generalize across domains or model families.
In contrast, \ourmethod measures sentiment distributional stability, a domain-agnostic property that emerges from LLMs’ tendency toward risk-averse, emotionally consistent outputs.
Our approach aligns with the latest advancements in behavioral modeling, extending the detection focus from local statistical anomalies to generalized, sentiment-invariant patterns.
Because it requires neither model access nor fine-tuning, our approach is well-suited for zero-resource, forensic, or black-box scenarios where transparency and generalizability are paramount.

\begin{figure*}[!t]
    \centering
    \includegraphics[width=\linewidth]{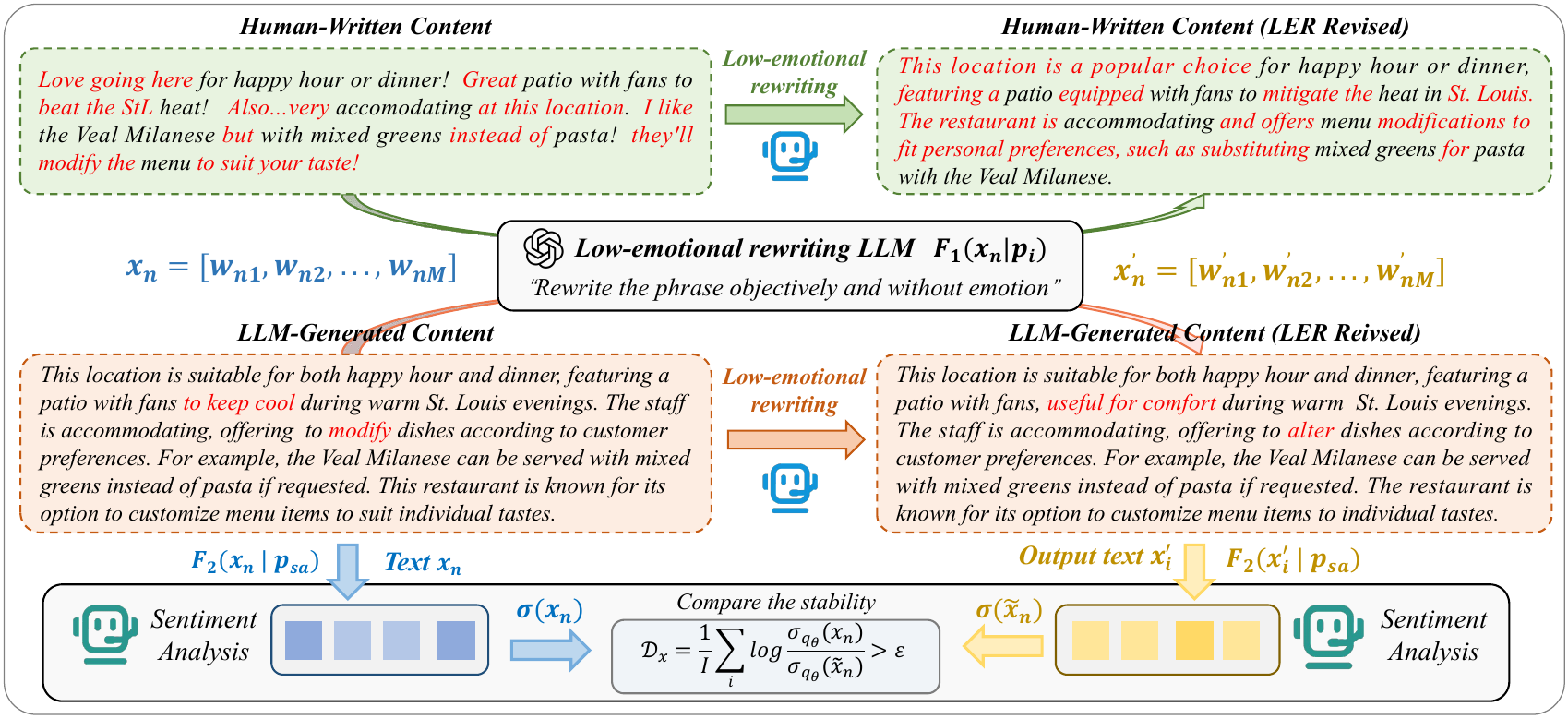}
    \caption{ Illustration of the proposed \ourmethod. Detecting LLM-generated text by sentiment distribution analysis through low-emotional stability rewriting. 
    See~\autoref{figure:low-emotional-rewrite-examples1} for more examples of low-emotional rewriting.}
    \label{figure:framework}
\end{figure*}

\section{Methodology}
\label{methodology}

\subsection{Threat Model and Problem Setup}
\label{threat-model-problem-setup}
LLM-generated texts can be misused to propagate misinformation and manipulate content; therefore, the goal of detection is to determine whether a given text is human-written or generated by an LLM under realistic forensic constraints. 
We consider a black-box setting, where the detector has access only to the final text, without relying on model, decoder probabilities, watermark signals, or task-specific supervision.
Formally, let \( \mathcal{P} \) and \( \mathcal{Q}_\theta \) denote the distributions of human-written and LLM-generated texts, respectively, over a metric space \( \mathcal{X} \). 
Given a set of candidate texts \( \{x_n\}_{n=1}^{N} \), where each \( x_n \in \mathcal{X} \) is drawn IID from either \( \mathcal{P} \) or \( \mathcal{Q}_\theta \), our goal is to determine its origin:
\begin{equation} 
    \hat{y}_n = \underset{ y_n \in \{y_\mathcal{P}, y_{\mathcal{Q}_\theta} \} }{\operatorname{argmax}} P (y_n \mid x_n, \mathcal{P}, \mathcal{Q}_\theta).
\end{equation}
Here, \( x_n = [w_{n1}, w_{n2}, \ldots, w_{nM}] \) denotes the \( n \)-th text sequence, and \( w_{nm} \) represents the \( m \)-th token in \( x_n \). 
When \( N = 1 \), the setting reduces to a single-text instance task~\cite{zhang2024detecting}. 
\autoref{table:notation} summarizes the key notation used throughout the paper.

Within this problem setting, we assume an adversary whose objective is to produce text that is indistinguishable from human writing while preserving semantics and task utility. 
The adversary has no access to detector parameters, internal embeddings, or decision thresholds, but may prompt or fine-tune an LLM and apply generic post-processing operations such as paraphrasing, style editing, or translation-based rewriting. 
Consequently, an effective detector should rely on behavioral properties that remain stable under such semantic-preserving transformations, rather than on fragile surface artifacts.

\subsection{Motivation and Detection Framework}
\label{motivation}
\textbf{Motivation.} Sentiment distributions exhibit stable differences between human-written and LLM-generated texts, even before any explicit transformation (\autoref{figure:scatter-plot}). 
Projecting texts into a sentiment-aware feature space reveals a consistent separation between the two sources, suggesting that sentiment patterns encode intrinsic behavioral cues that can be exploited for provenance attribution.
Further analysis shows that controlled neutralization of emotional tone amplifies the contrast: LLM-generated text tends to retain a relatively stable sentiment profile under low-emotional rewriting, whereas human-written text exhibits larger structural shifts. 
This phenomenon, referred to as \emph{emotional response inertia}, motivates the use of rewriting as a functional probe for detecting authorship.

Building on this insight, \ourmethodTT is introduced as a novel training-free, black-box detection framework. 
It quantifies the log-space divergence between sentiment features of the original text and its rewritten variants under carefully designed low-emotional rewriting. 
By analyzing changes in sentiment distributions induced by these controlled transformations, \ourmethodTT robustly discriminates between human-written and LLM-generated texts without requiring domain-specific supervision, watermarking, or white-box access.

\textbf{Overview of the detection pipeline.}
An overview of the proposed framework is shown in \autoref{figure:framework}. 
The pipeline consists of three stages:
\textbf{\textit{(I) Low-emotional rewriting}}: rewriting the input text with zero-shot LLM prompts designed to suppress overt emotional expression;
\textbf{\textit{(II) Sentiment feature extraction}}: extracting sentiment distribution features from both the original and rewritten texts; and
\textbf{\textit{(III) Sentiment distribution divergence analysis}}: measuring the divergence between these sentiment representations to identify stylistic stability and infer text origin.

\subsection{Analysis of Sentiment Distribution Feature in LLM-Generated Text}
In this subsection, we formalize the empirical observation introduced earlier by analyzing how sentiment expression patterns respond to controlled stylistic variation.
Building upon the distributional separation observed in ~\autoref{figure:scatter-plot}, we focus on the differential sensitivity of human and LLM-generated texts to low-emotional rewriting, which serves as the primary behavioral signal for discrimination.

\vspace{-0.3\baselineskip}
\begin{tcolorbox}[colback=gray!10, colframe=gray!90, boxrule=0.4pt, arc=1mm, left=4pt, right=4pt, top=4pt, bottom=4pt]
    \textbf{Key Observation: Emotional Response Inertia.} \\
    {LLM-generated texts largely preserve their inherent affective patterns after neutral-tone rewriting, whereas human-written texts show more pronounced and consistent shifts under the same controlled transformation.}   
\end{tcolorbox}
\vspace{-0.3\baselineskip}

\begin{table}[!t]
    \centering
    \small
    \caption{Main Notations and Explanations}
    \label{table:notation}
    \renewcommand{\arraystretch}{1} % row height
    \begin{tabular}{cp{0.73\linewidth}}
        \toprule
        \textbf{Notation} & \makecell[l]{\textbf{Explanation}} \\
        \midrule
        $x$ & Input text sample \\
        $x_i'$ & Rewritten text using low-emotional prompt $p_i$ \\
        $x^*$ & Twice-transformed text via inverse mapping: $\mathcal{F}^{-1}(\mathcal{F}(x_i'))$ \\
        $w_{i,j}$ & The $j$-th token in text $x_i$ \\
        $N$ & Number of rewriting prompts or samples \\
        $y$ & Indicator of text origin: human or LLM \\
        $q_\theta$ & LLM generating distribution \\
        $\mathcal{P}$ & Distribution of human-written texts \\
        $\mathcal{Q}_\theta$ & Distribution of LLM-generated texts \\
        $F_1(x \mid p_i)$ & Low-emotional rewriting function using $p_i$ \\
        $F_2(x \mid p_{sa})$ & Sentiment feature analysis function \\
        $p_i$ & Instruction for low-emotional rewriting \\
        $p_{sa}$ & Prompt for guiding sentiment analysis \\
        $\mathcal{F}, \mathcal{F}^{-1}$ & Inverse prompt pair for neutral semantic transformation \\
        $\sigma(x)$ & Sentiment feature vector of text $x$, $\sigma(x) \in \mathbb{R}^k$ \\
        $\text{SDC}(x)$ & Sentiment Distribution Consistency score \\
        $\text{SDP}(x)$ & Sentiment Distribution Preservation score \\
        $\mathcal{D}_x$ & Sentiment divergence score for classification \\
        $\varepsilon$ & Threshold for detection decision \\
        $\hat{y}$ & Predicted label of text source \\
        \bottomrule
        \end{tabular}
\end{table}

This measurable divergence originates from the fundamentally different ways in which humans and LLMs encode sentiment across stylistic variation. 
To systematically characterize this divergence, we adopt a two-stage sentiment analysis framework that explicitly decouples affective suppression through rewriting from distributional characterization via feature extraction.
First, we define a transformation function \( F_1(x_n \mid p_i) \) that applies a low-emotional rewriting prompt \( p_i \) to produce a rewritten sample \( x' \). 
To be specific, $F_1(\cdot \mid p_i)$ maps text $x_n$ to $x^{\prime}_{n}= [w_{n1}^{\prime}, w_{n2}^{\prime}, \ldots, w_{nJ}^{\prime}]$ of indefinite length.
This rewriting process is repeated using a set of prompts \( \{p_i\}_{i=1}^I \), yielding a collection \( \{x_n'\}_{n=1}^N \).

We denote the sentiment distribution of a text by a function \( \sigma(x) \in \mathbb{R}^k \), which maps input text to a sentiment feature vector based on \( k \)-dimensional psychological or semantic components (e.g., valence, arousal, polarity).
Formally, we analyze sentiment shifts as  \( \sigma(x_n) = F_2(x_n \mid p_{sa}) \), and \( F_2(\cdot \mid p_{sa}) \) denotes a sentiment analysis function guided by prompt \( p_{sa} \), and each vector \( \sigma(\cdot) \) captures the underlying emotional distribution of the text.  
By comparing the two sentiment feature vectors, we expose structural variability characteristic of the text’s origin.
We elaborate on these two stages below.
\begin{figure*}[!th]
    \centering
    \includegraphics[width=\linewidth]{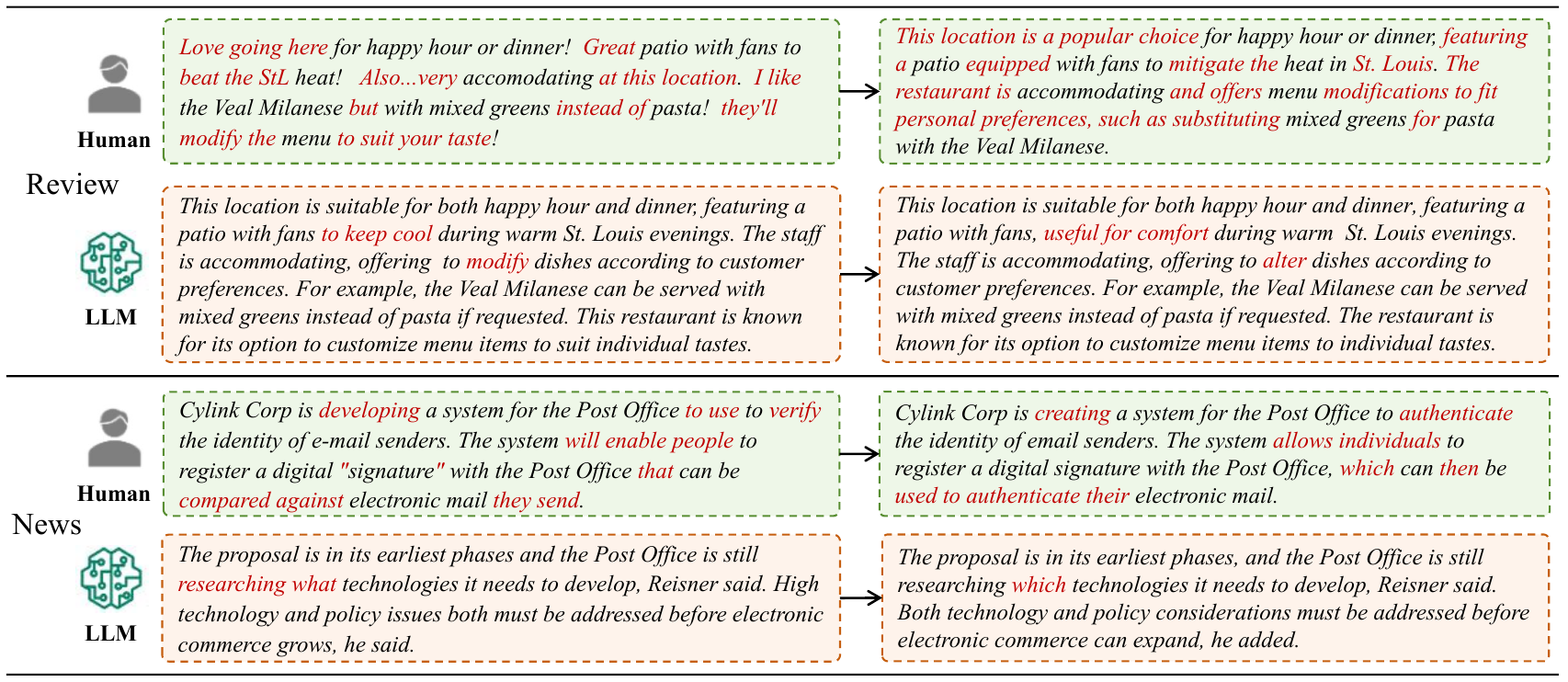}
    %fig5.pdf}
    \caption{Illustration of low-emotional rewriting examples on \datasetreview dataset and \datasetnews dataset for \ourmethod.
    The light red section represents the text that has been modified by low-emotional rewriting.}
    \label{figure:low-emotional-rewrite-examples1}
\end{figure*}

\subsubsection*{\textbf{Low-Emotional Rewriting (LER)}}
Given an input sequence \( x = [w_1, w_2, \ldots, w_M] \), we define a low-emotional rewriting function \( F_1(x \mid p_i) \) that transforms \( x \) into a stylistically neutral form under instruction prompt \( p_i \). This transformation preserves the core semantics while attenuating subjective or emotional expressions.  
In practice, \( F_1 \) is realized via zero-shot prompting of an LLM with low-emotion style instructions:
\begin{itemize}
    \item \textit{``Rewrite this more straightforwardly.''}
    \item \textit{``Polish this into a machine-like objective tone.''}
\end{itemize}
 We employ a diverse set of prompts \( \{p_i\}_{i=1}^{I} \) to generate a corresponding set of rewritten outputs \( \{x_n^{\prime}\}_{n=1}^{N} \).
\autoref{figure:low-emotional-rewrite-examples1} showcases LER examples on review and news domains.

In emotionally rich contexts, human-written texts tend to undergo substantial sentiment changes, whereas LLM-generated texts generally show more stable sentiment patterns. 
This phenomenon is also observed in other types of texts, and will be further quantified and illustrated in the experimental results section. 
Accordingly, our method does not rely on the magnitude of expressed emotion or domain-specific affective keywords. 
Instead, it measures distributional changes in sentiment features between the original and rewritten texts in log space, where LLM-generated samples tend to exhibit higher rigidity under controlled rewriting than human-written samples.

\subsubsection*{\textbf{Sentiment Feature Analysis}}
Following the low-emotional rewriting process, we quantify the sentiment characteristics of both the original and rewritten texts using a sentiment analysis function \( F_2(\cdot \mid p_{sa}) \). 
As illustrated in \autoref{figure:framework}, it is utilized to quantify the sentiment feature of the text $x$ and the rewritten samples \( x^{\prime} = F_1(x \mid p_i) \): 
\begin{equation}
    \sigma(x) = F_2(x \mid p_{sa}) \in \mathbb{R}^k,
\end{equation}
\begin{equation}
    \sigma(x^{\prime}) = F_2(x^{\prime} \mid p_{sa}) = F_2 (F_1(x \mid p_i) \mid p_{sa}).
\end{equation}
where the sentiment analysis prompt \( p_{sa} \) is designed to elicit detailed sentiment components (e.g., valence, polarity).

\subsection{Detection via Sentiment Distribution Divergence} 
With the help of the concepts of sentiment analysis functions $F_1(x \mid p_i)$ and $F_2(x^{\prime} \mid p_{sa})$, we can define the distribution divergence in sentiment feature:
\begin{equation}
\begin{aligned}     
\mathcal{D}(x, q_\theta) & \triangleq \log \sigma(x) - \mathbb{E}_{x^{\prime} \sim F_1(x \mid p_i)} \log \sigma(x^{\prime}),
\end{aligned} 
\end{equation}
where $\sigma(x) = F_2 (x \mid p_{sa})$.
We formally refer to this as the sentiment distribution divergence hypothesis, which describes characteristic differences between LLM-generated and human-written texts.
For the text $x$ generated from LLM $q_\theta$, $\mathcal{D}(x, q_\theta)$ typically shows smaller values. 
For human-written text $x$, $\mathcal{D}(x, q_\theta)$ yields larger values. 
Furthermore, the two can be distinguished by the aggregated change of sentiment features:
\begin{equation}    
\mathcal{D}_x =\frac{1}{I} \sum\limits_i \log \left(\frac{\sigma_{q_\theta}(x)}{\sigma_{q_\theta}(F_1(x \mid p_i))}\right)
\end{equation}
Let $\varepsilon$ be a fixed decision threshold. We identify the input text $x$ as LLM-generated when $\mathcal{D}_x < \varepsilon$.

% where $\varepsilon$ is a predefined threshold value, with the input text $x$ identified as LLM-generated when the divergence $\mathcal{D}_x < \varepsilon$.  

Then, we formalize our detection strategy by introducing two metrics that capture the unique stability of LLM-generated text: the basic \textit{Sentiment Distribution Consistency} and the variant \textit{Sentiment Distribution Preservation}.
These metrics form the basis of an end-to-end, threshold-based decision rule.

\subsubsection*{\textbf{Sentiment Distribution Consistency}}
We define SDC to quantify the degree to which the sentiment distribution of a given text remains stable after low-emotional rewriting.
The underlying intuition is that LLM-generated texts, shaped by autoregressive nature and statistical regularities in training data, exhibit greater invariance in sentiment distribution under style-neutral prompts.

Given an input text $x$, and rewriting prompts $\{p_i\}_{i=1}^I$, we first generate rewritten texts $x' = F_1(x \mid p_i)$. 
Using a zero-shot sentiment feature extractor $F_2(\cdot \mid p_{sa})$, we compute the sentiment vectors $\sigma(x)$ and $\sigma(x')$. 
We then quantify the sentiment distribution consistency through:
\begin{equation}
    L_1 = || \log F_2(x | p_{sa}), \log F_2(F_1 (x | p_i) | p_{sa})||.
\end{equation}

where \( \sigma(x) = F_2(x \mid p_{sa}) \in \mathbb{R}^k \) is the sentiment feature extracted via zero-shot prompting.  
If the text $x$ is generated from LLM, \textit{it often shows smaller SDC value}.
The transformation prompts $p_i$ used in LER include examples such as: 
\begin{itemize}[itemsep=1pt, topsep=1.5pt]
    \item \textit{Rewrite this more straightforwardly.}
    \item \textit{Polish this in a machine-like objective tone.}
    \item \textit{Rewrite this objectively and without emotion.}
\end{itemize}
These can be optimized via automatic prompt selection methods~\cite{autoprompt:ma2024fairness}.

\subsubsection*{\textbf{Sentiment Distribution Preservation}}
We further define SDP to evaluate sentiment-invariant patterns under a pair of semantically neutral, invertible transformations~\cite{baseline3:RAIDAR:mao2024detecting}. 
This property is based on the observation that LLM-generated texts tend to retain sentiment structure even when edited in ways that do not affect semantics (e.g., paraphrasing, abbreviation expansion, or rephrasing).  
Suppose we perform sentiment-independent and inverse transformations (such as converting people or extensions and abbreviations) on the texts. 
In that case, it will \textit{output with the same sentiment distribution as the input text in the previous}.
We refer to the property that model-generated text maintains stable sentiment distributions under semantic-preserving transformations as sentiment distribution preservation.

Let $\mathcal{F}$ and $\mathcal{F}^{-1}$ be a pair of inverse prompts (e.g., ``expand this paragraph'' and shorten this paragraph''). 
We define the SDP score by transforming $x$ into $x^*$, then measuring the sentiment difference:
\begin{equation}
% \fontsize{10}{12}\selectfont
    L_2 = || \log F_2(\mathcal{F}^{-1}( \mathcal{F}(F_{1}(x) | p_i) ) | p_{sa} ), \log F_2(x | p_{sa})||.
\end{equation}
where
\begin{equation}
    x^* = \mathcal{F}^{-1}(\mathcal{F}(F_1(x \mid p_i))).
\end{equation}
The lower SDP score indicates higher preservation of sentiment under inverse transformations, which is a characteristic commonly found in LLM-generated content.
By mapping twice through LLM instructions with opposite meanings, the LLM-generated text preserves its original sentiment distribution feature.
This metric captures a deeper property: even after multiple stylistic but semantically neutral transformations, LLM-generated texts maintain their sentiment profiles more rigidly than human-written texts.

\begin{algorithm}[!t]
    \caption{\ourmethod LLM-Generated Text Detection}
    \label{algo1}
    \begin{algorithmic}[1]
    \STATE \textbf{Input}: text $x$, prompt list $\{p_i\}$, analysis prompt $p_{sa}$, inverse transforms $\mathcal{F}, \mathcal{F}^{-1}$, threshold $\varepsilon$.
    \STATE \textbf{Output}: Prediction $\hat{y}$ of the input text $x$.
     \STATE Design elaborate instructions $p_1, p_2, \ldots, p_I$
    \STATE Design sentiment feature analysis instruction $p_{sa}$
    \STATE Compute the original sentiment $\sigma(x) = F_2(x \mid p_{sa})$
    \FOR{$i$ in $[1, I]$}
        \STATE Generate LER rewritten text $x^{\prime} = F_1(x | p_i)$
        \STATE (Optional) Generate $x^* = \mathcal{F}^{-1}(\mathcal{F}(x'))$ and $\sigma(x^*)$
        \STATE Conduct sentiment analysis $\sigma(x') = F_2(x' \mid p_{sa})$
        \STATE Calculate sentiment distribution consistency $L_1^i$:
        $$ L_1 = || \log F_2(x | p_{sa}), \log F_2(F_1 (x | p_i) | p_{sa})|| $$    
        or sentiment distribution preservation $L_2^i$:
        $$L_2 = || \log F_2(\mathcal{F}^{-1}( \mathcal{F}(F_{1}(x) | p_i) ) | p_{sa} ), \log F_2(x | p_{sa})||$$
        \STATE Compute $L_1^i = |\log \sigma(x) - \log \sigma(x')|_1$, \\ 
        $L_2^i = |\log \sigma(x) - \log \sigma(x^*)|_1$
    \ENDFOR
    \STATE Calculate $\text{SDC}(x) = \frac{1}{I}\sum\limits_i L_1$, $\text{SDP}(x) = \frac{1}{I}\sum\limits_i L_2$
    \STATE Compute divergence score $\mathcal{D}_x = \text{SDC}(x) \text{ or } \text{SDP}(x)$
    \RETURN $\mathbb{I}[\mathcal{D}_x < \varepsilon]$
\end{algorithmic}
\end{algorithm}

In the implementation of low-emotional equivariance, instead of directly low-emotional rewriting texts, we apply a set of mutually inverse transformations to the text under evaluation.
These transformations are independent of sentiment (such as changing the person, extensions and abbreviations).
If the output text exhibits the same emotional intensity before and after transformation, we refer to this property as low-emotional equivariance. 
We denote these mappings as $\mathcal{F}$ and $\mathcal{F}^{-1}$.

Here, $\mathcal{F}$ represents the transformation prompt used to request LLM-generated output with the transformation applied, while $\mathcal{F}^{-1}$ represents the prompt with the opposite transformation.
We provide specific examples of these prompt designs:
\begin{itemize}
    \item Positive transformation $\mathcal{F}$: \textit{Please rewrite this in the third person while maintaining its meaning.}
    \item Inverse transformation $\mathcal{F}^{-1}$: \textit{Please rewrite this in the first person while maintaining its meaning.}
    \item Positive transformation $\mathcal{F}$: \textit{Please expand this paragraph without changing its meaning.}
    \item Inverse transformation $\mathcal{F}^{-1}$: \textit{Please abbreviate this paragraph without changing its meaning.}
\end{itemize}
In summary, through these two opposite-meaning prompt mappings, the LLM-generated can restore the initial stable low-emotional intensity of the text.

Algorithm~\autoref{algo1} describes the full detection process, combining rewriting, sentiment analysis, and classification in an end-to-end fashion. 
% Detailed prompt examples and rewriting outputs are presented in Section~\ref{sec:experiments}.}
It takes as inputs the raw text sample $x$, the source language model $q_\theta$, a sentiment transformation function $F_1(x|p_i)$, a sentiment distribution analyzer $F_2(x^{\prime}|p_{sa})$, and a decision threshold $\varepsilon$. 
The detection process begins by designing two distinct types of instructions: (1) a set of $I$ elaborate text rewriting prompts $p_1, \ldots, p_I$ that guide the modification of input text, and (2) a specialized sentiment analysis prompt $p_{sa}$ for sentiment feature extraction. 
For each rewriting prompt $p_i$, the algorithm first generates the transformed text $x^{\prime} = F_1(x|p_i)$ through the sentiment transformation function, then extracts the corresponding sentiment features using $F_2(x^{\prime}|p_{sa})$. 
The core detection metric $\mathcal{D}_x$ is computed through one of two complementary approaches: either evaluating the sentiment distribution consistency, which measures the divergence between original and rewritten text sentiment profiles, or assessing the sentiment distribution preservation, which quantifies the feature stability under inverse transformations. 
The algorithm then aggregates these measurements across all rewritten texts to compute the final divergence score $\mathcal{D}_x$, ultimately returning the detection result based on the threshold comparison $\mathcal{D}_x < \varepsilon$, where True indicates LLM-generated content and False denotes human-written text.

\begin{figure*}[!th]
    \centering
    \includegraphics[width=\linewidth]{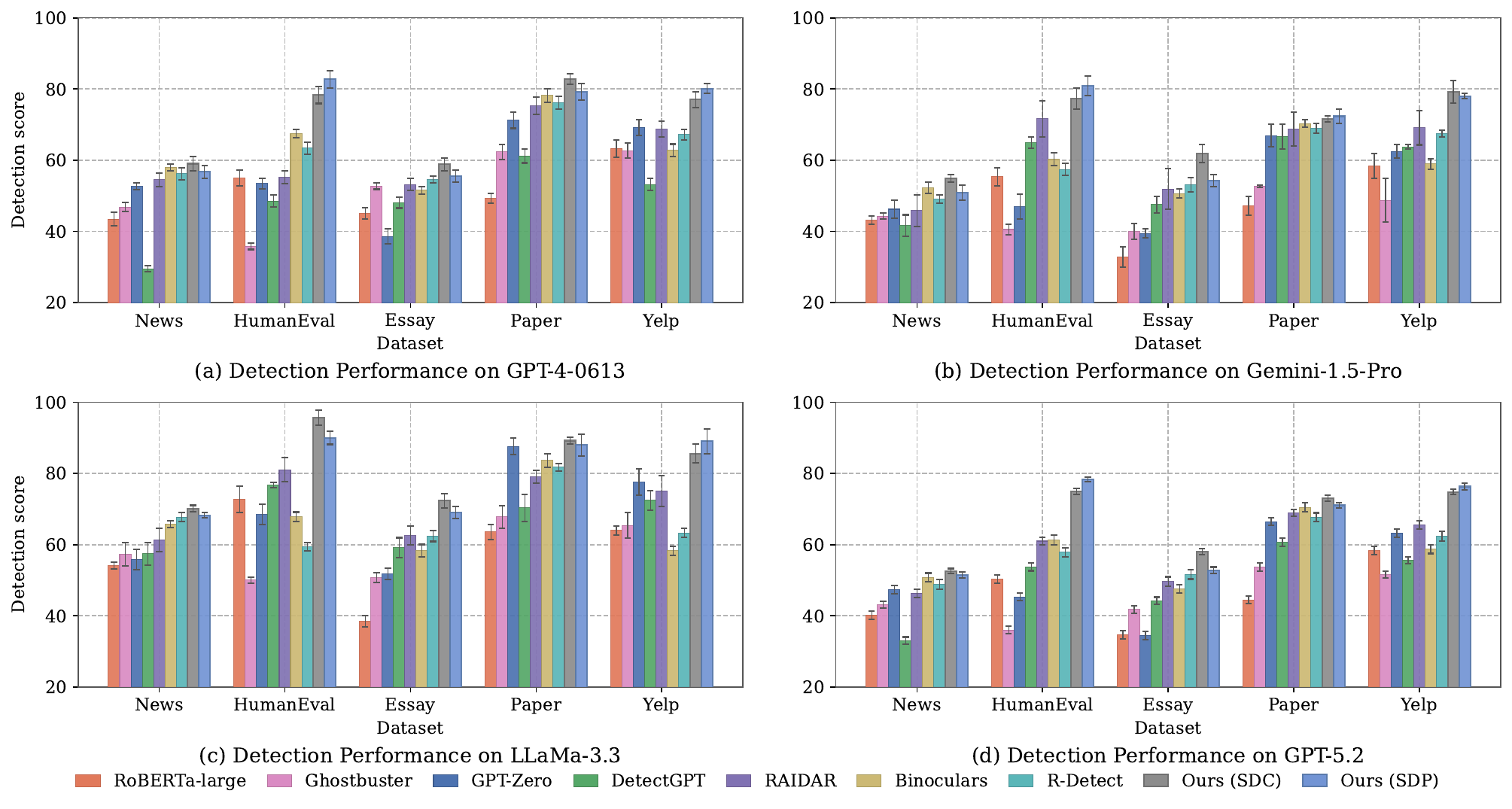}
    \caption{ Main detection F1 score on texts generated by \GPTFourAll, \GeminiOneFivePro, \LLaMa, and \GPTFiveTwoAll across five domains, including \datasetnews, \datasetcode, \datasetessay, \datasetpaper, and \datasetreview.
    \ourmethod consistently achieves the best overall detection performance over all competing baselines.}
    \label{figure:main-detection}
\end{figure*}

\section{Experiments}
\subsection{Experimental Setup}

\subsubsection{Datasets}
We evaluate five text corpora covering diverse text generation scenarios, with the goal of examining behavioral differences in sentiment distribution between LLM-generated and human-written texts. 
We utilize these datasets to probe the stability of emotional tone across domains, styles, and content structures.

\textbf{\datasetnews Dataset}~\cite{verma2024ghostbuster} consists of 5,000 real news articles from 50 journalists, with corresponding machine-generated versions created by \GPTThree using a two-step process: generating titles, then full articles.
It captures the neutral tone of factual reporting, enabling analysis of how affective cues persist or diminish across rewrites.

\textbf{\datasetcode Dataset}~\cite{HumanEvalaCode:chen2021evaluating} contains 164 handwritten programming problems with function signatures, docstrings, and test cases covering various programming concepts. 
Though not overtly emotional, stylistic patterns in comments and naming provide indirect sentiment signals. This domain tests whether sentiment-preserving structure emerges even in functional content.

\textbf{\datasetessay Dataset}~\cite{verma2024ghostbuster} is a student paper dataset based on the British Academic Written English corpus, including high school and university-level papers from various disciplines. 
These essays provide a rich testbed for exploring emotional trajectory variation across academic discourse, with stylistic complexity and argument structure variation.

\textbf{\datasetreview Dataset}~\cite{baseline3:RAIDAR:mao2024detecting} is a modified test dataset based on the Yelp review dataset. 
We extracted 1,000 human-written reviews and LLM-generated reviews with matched length.

\textbf{\datasetpaper Dataset} contains 500 abstracts from recent publications (ACL 2023–2024). 
We use \GPTThreeFiveTurbo to regenerate each abstract based on the title and the first 15 words of the original content. 
By including only post-2023 papers, we design the evaluation to follow a zero-shot prompting setting.

\subsubsection{Baselines}
We compare \ourmethod with state-of-the-art text provenance detection methods:

\textbf{GPTZero}~\cite{GPTZero} is a commercial classifier relying on handcrafted features and syntactic heuristics for detecting LLM-generated content.
Its performance serves as a strong practical baseline for AI detection, capable of detecting \textit{ChatGPT}, \textit{LLaMa}, and other LLMs.

\textbf{Ghostbuster}~\cite{verma2024ghostbuster} performs detection via direct language models querying, without the need for any additional training, fine-tuning, or labeled data to adapt to specific domains or model types. 

\textbf{DetectGPT}~\cite{DetectGPT:mitchell2023detectgpt} identifies LLM-generated text by examining changes in curvature of log-probability under small input perturbations.
Unlike the black-box approach, it relies on model scoring and assumes internal access to an LLM.

\textbf{Fast-DetectGPT}~\cite{bao2024fast} improves upon DetectGPT by substituting the perturbation step with a conditional probability curvature analysis derived from efficient sampling, allowing for faster and more accurate zero-shot detection.

\textbf{Binoculars}~\cite{baseline2:Binoculars:hans2024spotting} introduces a zero-shot detection framework that contrasts two different language models, which compute a score based on the ratio of the observer's perplexity to its cross-perplexity with the performer.

\textbf{RAIDAR}~\cite{baseline3:RAIDAR:mao2024detecting} analyzes text modification patterns to amplify model-specific signals through rewriting-based contrastive features. 
It generates multiple paraphrased variants of each input and compares their responses to detect subtle divergences characteristic of LLM-generated content. 

\textbf{R-Detect}~\cite{baseline1:R-Detect:song2025deep} is a distribution-based baseline that detects LLM-generated text by computing divergence metrics across multiple candidate rewrites. 
It captures structural regularities without requiring access to model internals or probabilities.

\subsubsection{Implementation Details}
In our experiments, we first determine a suitable threshold value by evaluating performance on a dedicated validation set, and once chosen, we maintain this threshold consistently across all evaluation tasks to ensure comparability. 
For each input sample, we produce $N=5$ semantically equivalent variants using a predefined set of sentiment-invariant rewriting prompts (e.g., paraphrasing, person shifting, length variation). 
The rewriting is performed with \GPTFourAll, employing a fixed decoding configuration with temperature $=0.7$ and top-p $=0.9$. 
These parameters were chosen to balance variability and coherence.
All rewriting operations use the same model and decoding parameters consistently across all domains, ensuring that variability in outputs is caused solely by the prompts.

\subsection{Main Detection Performance}
This section presents the experimental validation of the proposed \ourmethodTT developed in Section~\ref{methodology}. 
To evaluate the general detection effectiveness and model-agnostic adaptability of our proposed framework, we conduct experiments on both commercial and open-source LLMs, covering a spectrum of architectures and training pipelines. 
We evaluate \ourmethodTT across five representative domains, including news, code, student essays, academic abstracts, and user reviews, using \textit{\ourmethod-SDC} and \textit{\ourmethod-SDP} as two core variants.

\textbf{\textit{(I) Main Results Across Advanced LLMs.}} As illustrated in \autoref{figure:main-detection}, \ourmethod, including both the SDC and SDP variants, consistently outperforms all considered baselines across the four advanced LLMs evaluated in this study: \GPTFourAll, \GeminiOneFivePro, \LLaMa, and \GPTFiveTwoAll.
Notably, while strong baselines like Binoculars and R-Detect retain competitive performance on certain specific subsets of the datasets, their effectiveness diminishes substantially when applied to the more advanced language models or across heterogeneous domains, with particularly pronounced performance drops observed in news article and essay detection tasks. 
This pattern indicates that the stylistic fluency, lexical diversity, and coherence of modern LLM-generated outputs increasingly challenge existing detectors that rely on superficial or artifact-dependent cues.

In contrast, \ourmethod demonstrates a consistently high level of F1 scores across all four LLM models and five diverse domains, encompassing both closed-source commercial models and open-source models, highlighting its broad applicability and robustness. 
Specifically, it achieves F1 scores exceeding 80\% on code and review detection tasks for \GPTFourAll and \GeminiOneFivePro, while maintaining strong performance above 75\% in detecting academic paper texts. 
The academic paper domain, characterized by its formal writing style, technical vocabulary, and structured distributional patterns, represents a particularly difficult challenge for traditional detectors.
Importantly, \ourmethod preserves a clear performance advantage over all baselines in this domain, evidencing not only its effectiveness under standard conditions but also its robustness to domain shifts and stylistic variations inherent in academic writing. 
This suggests that the method’s underlying sentiment-invariant divergence analysis provides more generalizable signals.

\begin{figure}[!th]
    \centering
    \includegraphics[width=\linewidth]{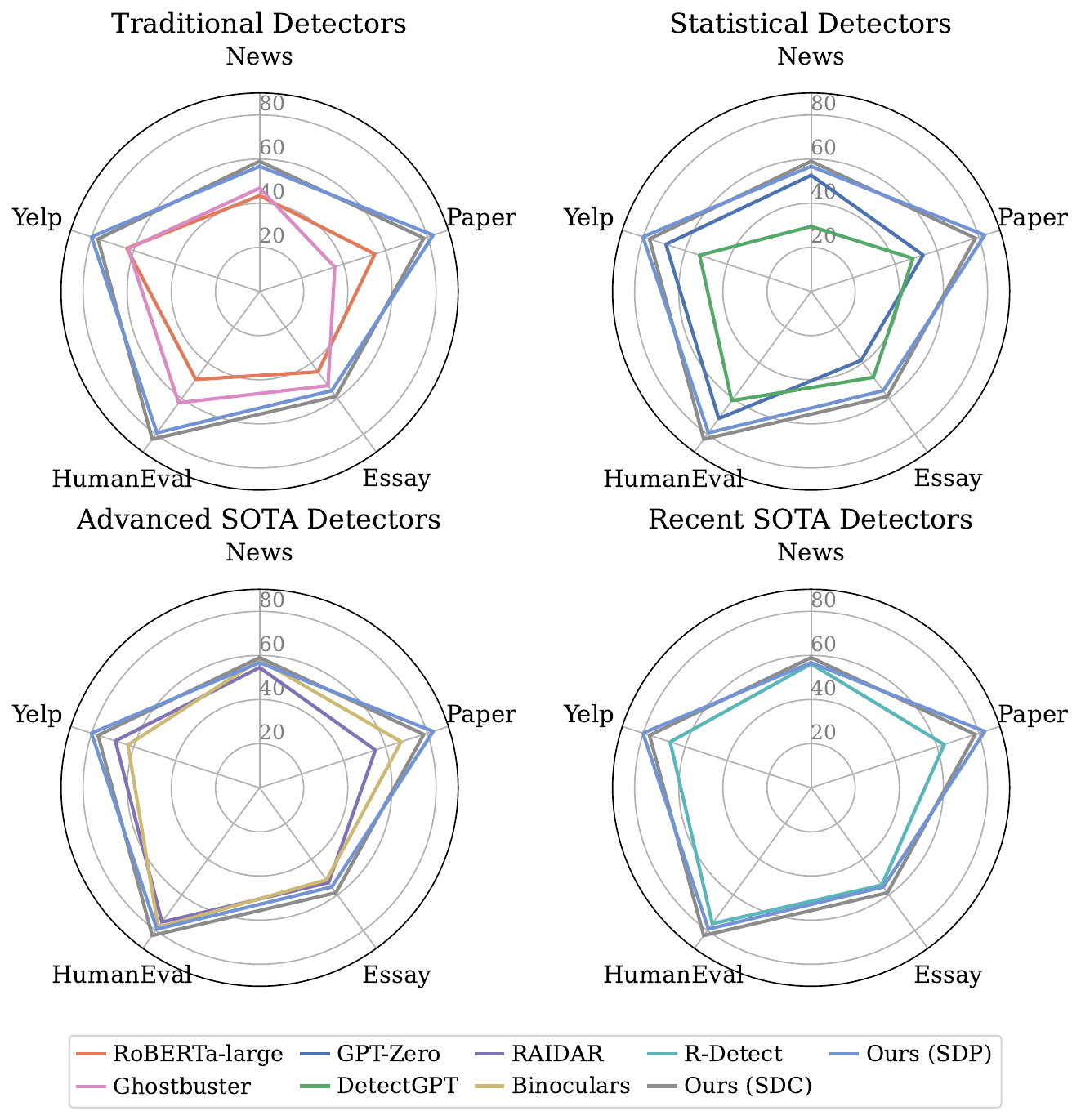}
    \caption{Comparative results of the performance of \ourmethod against four types of approaches on \GPTFourAll, including traditional methods, OpenAI RoBERTa, business detectors, and SOTA detectors.}
    \label{figure:datasets}
\end{figure}

\textbf{\textit{(II) Performance Across Diverse Text Domains.}}
As demonstrated in \autoref{figure:datasets}, \ourmethod (SDC/SDP) exhibits comprehensive advantages over all evaluated baselines, including RoBERTa-large, Ghostbuster, GPT-Zero, DetectGPT, RAIDAR, Binoculars, and R-Detect, across all five evaluated domains (\datasetnews, \datasetpaper, \datasetessay, \datasetcode, and Yelp).
The experimental results reveal particularly remarkable performance gaps in code and academic paper detection, two domains that pose significant challenges for existing detectors due to their specialized stylistic and structural characteristics.
For HumanEval code detection and academic paper identification, \ourmethod achieves substantial leads of at least 10 percentage points in F1 scores compared to the second-best baselines (Binoculars and R-Detect), confirming the superior effectiveness of sentiment-invariant features in capturing artifacts of specialized textual content.
Besides, in the challenging \datasetessay domain, \ourmethod maintains a clear advantage over all baselines, outperforming even state-of-the-art methods like Binoculars and RAIDAR by a notable margin.
While the performance differences are relatively smaller for news articles and Yelp reviews, where baselines such as GPT-Zero and RAIDAR perform comparatively well, \ourmethod still secures stable superiority with approximately 5-7 percentage point margins.
This consistent and significant outperformance across five fundamental text categories provides compelling evidence that \ourmethod possesses universal discriminative capabilities that transcend traditional text-type boundaries.
As visually corroborated by \autoref{figure:datasets}, the radar chart results strongly support our methodology's effectiveness in diverse scenarios.
These findings suggest our technique captures more fundamental and generalizable detection signals than conventional model-based or statistical baselines, enabling robust performance across varied text domains.

\textbf{\textbf{(III) Detection Across Evolving Series Models.}}
\begin{figure}[!t]
    \centering
    \includegraphics[width=\linewidth]{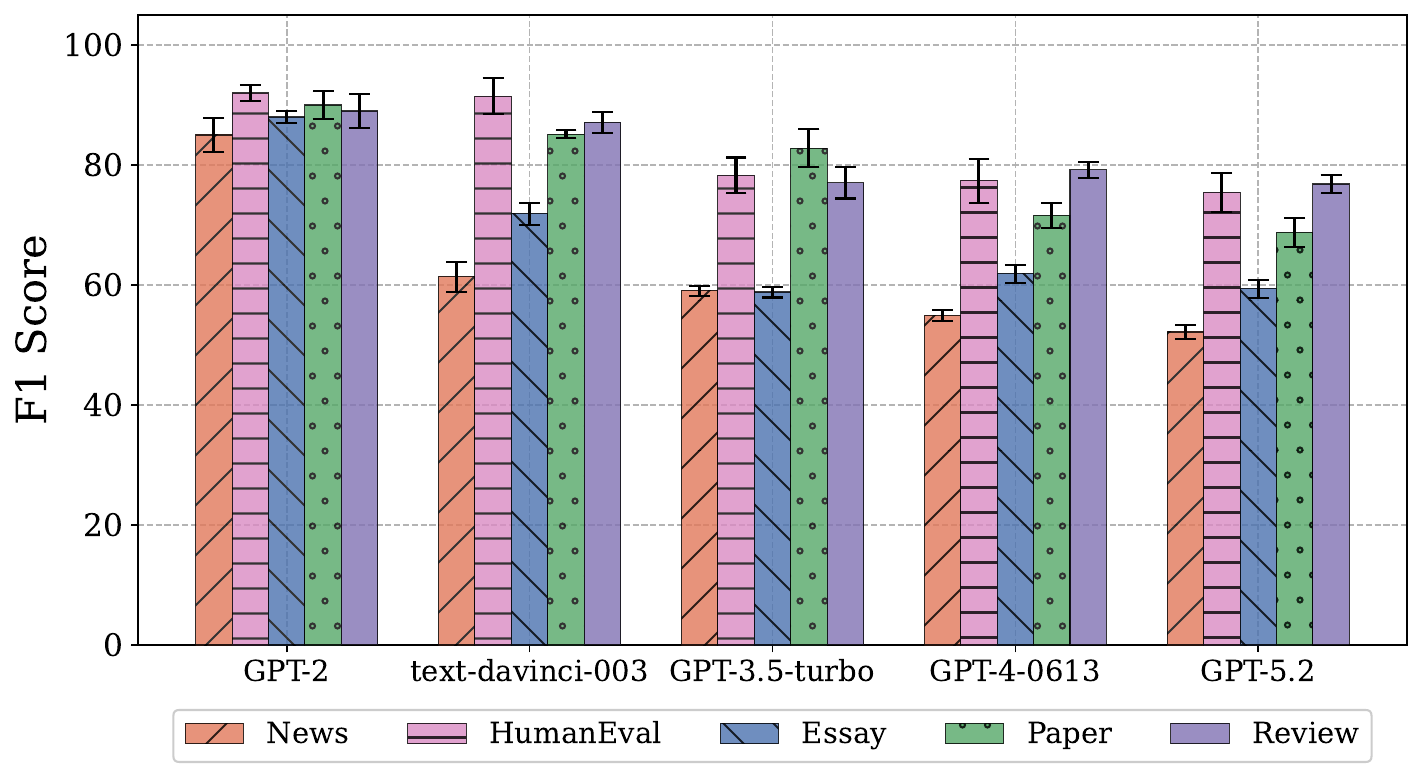}
    \caption{Comparative analysis of \ourmethod performance across GPT model generations: Demonstrating superior detection accuracy on \GPTThree, \GPTThreeFiveTurbo, \GPTFourAll, and \GPTFiveTwoAll in \datasetnews, \datasetcode, and academic text domains.}
    \label{figure:gptseries}
\end{figure}
At the same time, \ourmethod consistently outperforms baseline detectors across five GPT generations from GPT-2 to \GPTFiveTwoAll and five text domains, as shown in~\autoref{figure:gptseries}.
It maintains high accuracy ranging from 85\% to 92\% on \GPTFourAll and \GPTFiveTwoAll outputs, while baseline methods drop to between 60\% and 75\%, underscoring its robustness as models grow more sophisticated.
This trend is particularly important because later GPT-series models generally produce more fluent, better-aligned, and stylistically refined outputs, which reduce the effectiveness of detectors that rely primarily on superficial statistical irregularities or model-specific artifacts.
Besides, rather than showing a monotonic improvement or decline, the results reveal that the effect of model evolution is domain-dependent: detection remains relatively stable in some domains, while becoming more variable in others as generation quality improves.
This indicates that model progression does not affect all textual domains uniformly and that the impact of newer generations should be analyzed together with domain.

A closer look at the figure shows that \datasetcode, \datasetpaper, and \datasetreview remain comparatively stable across generations, with only limited variation from earlier GPT models to \GPTFourAll and \GPTFiveTwoAll. 
By contrast, \datasetnews and \datasetessay exhibit more noticeable fluctuations, suggesting that these domains are more sensitive to stylistic refinement introduced by newer GPT models.
In particular, the separation becomes less regular in the middle generations but recovers on later ones, implying that the challenge is not simply “newer models are harder,” but that different generations may alter different aspects of textual behavior.
Overall, the figure highlights that the proposed signal tracks cross-generation variation in a relatively controlled manner.
And the increasing gap across GPT versions suggests that \ourmethod is better suited to handle ongoing LLM evolution.

\vspace{-0.2\baselineskip}
\begin{tcolorbox}[colback=gray!10, colframe=gray!90, boxrule=0.4pt, arc=1mm, left=4pt, right=4pt, top=4pt, bottom=4pt]
    \textbf{Finding 1: Cross-Model Robustness and Generality.} \\ 
    (1) \ourmethod remains consistently effective across advanced LLMs from different model families. \\
    (2) Its advantage is particularly clear in challenging domains such as code and academic papers, and remains stable under GPT-series evolution.
\end{tcolorbox}
\vspace{-0.2\baselineskip}

\subsection{Sentiment-Invariant Patterns Analysis}
\begin{table}[!t]
    \centering
    \small
    \caption{Standard deviation of F1 scores across five domains. Lower values indicate better detection stability. 
    \ourmethod achieve the most consistent performance.}
    \label{table:std}
    \setlength{\tabcolsep}{4pt}
    \resizebox{\linewidth}{!}{
    \begin{tabular}{l||ccccc|c}
        \toprule
        \textbf{Method} & \textbf{Reuter} & \textbf{HumanEval} & \textbf{Essay} & \textbf{Paper} & \textbf{Yelp} & \textbf{Std.} \\
        \midrule
        LogRank & 4.85 & 6.21 & 5.74 & 5.66 & 6.12 & 5.72 \\
        \rowcolor{gray!10} RoBERTa-base & 6.03 & 4.92 & 6.18 & 5.79 & 6.37 & 5.86 \\
        RoBERTa-large & 5.64 & 5.88 & 6.92 & 6.22 & 6.44 & 6.02 \\
        \rowcolor{gray!10} Ghostbuster & 5.93 & 6.05 & 6.34 & 5.87 & 6.81 & 6.20 \\
        GPT-Zero & 6.89 & 5.53 & 7.15 & 6.74 & 8.27 & 6.92 \\
        \rowcolor{gray!10} DetectGPT & 6.74 & 5.97 & 6.83 & 6.61 & 7.11 & 6.65 \\
        Fast-DetectGPT & 4.76 & 5.21 & 4.93 & 5.17 & 5.41 & 5.10 \\
        \rowcolor{gray!10} Binoculars & 4.42 & 5.09 & 4.81 & 5.03 & 5.32 & 4.93 \\
        RAIDAR & 6.21 & 5.42 & 6.17 & 6.44 & 7.06 & 6.26 \\    
        \rowcolor{gray!10} R-Detect & 4.15 & 4.93 & 4.52 & 4.77 & 5.04 & 4.73 \\
        \midrule
        \ourmethod (SDC) & 4.38 & 5.16 & 4.71 & 5.09 & 5.25 & \textbf{4.72} \\
        \rowcolor{gray!10} \ourmethod (SDP) & 3.91 & 4.87 & 4.26 & 4.68 & 5.11 & \textbf{4.36} \\
        \bottomrule
    \end{tabular} }
\end{table}

\textbf{\textit{(I) Cross-domain Stability.}}
Beyond absolute detection accuracy, a practical detector should also maintain stable performance across heterogeneous text domains.
To evaluate this property, we report the standard deviation of F1 scores across five representative domains, including \datasetnews, \datasetcode, \datasetessay, \datasetpaper, and \datasetreview.
As shown in~\autoref{table:std}, \ourmethodTT achieves the lowest cross-domain variation among all compared methods.
Compared with strong recent baselines such as Binoculars, Fast-DetectGPT, and R-Detect, the fluctuation of \ourmethod is consistently smaller, and the reduction is even more evident relative to earlier baselines such as GPTZero, DetectGPT, and RAIDAR.
Overall, this means that the sentiment-invariant signal is less sensitive to domain-specific writing conventions and remains more stable when the text type changes.

The same trend is reflected in domains with stronger stylistic or structural constraints.
In particular, \ourmethod remains consistently competitive in \datasetcode and \datasetpaper, where many detectors tend to become less reliable because the surface form is more regular and the available stylistic cues are more limited.
A similar pattern is observed in \datasetessay, where broader stylistic diversity often increases the variance of competing methods.
Taken together, these results suggest that the proposed signal is not effective only in a few favorable cases, but remains comparatively insensitive to shifts in topic, formality, and textual structure.
This property is especially desirable in provenance attribution scenarios, where the detector is expected to operate across mixed domains without domain-specific recalibration.

\textbf{\textit{(II) Sentiment Stability as an Interpretable Signal for LLM-Generated Content.}}
\begin{figure}[!t]
    \centering
    \includegraphics[width=\linewidth]{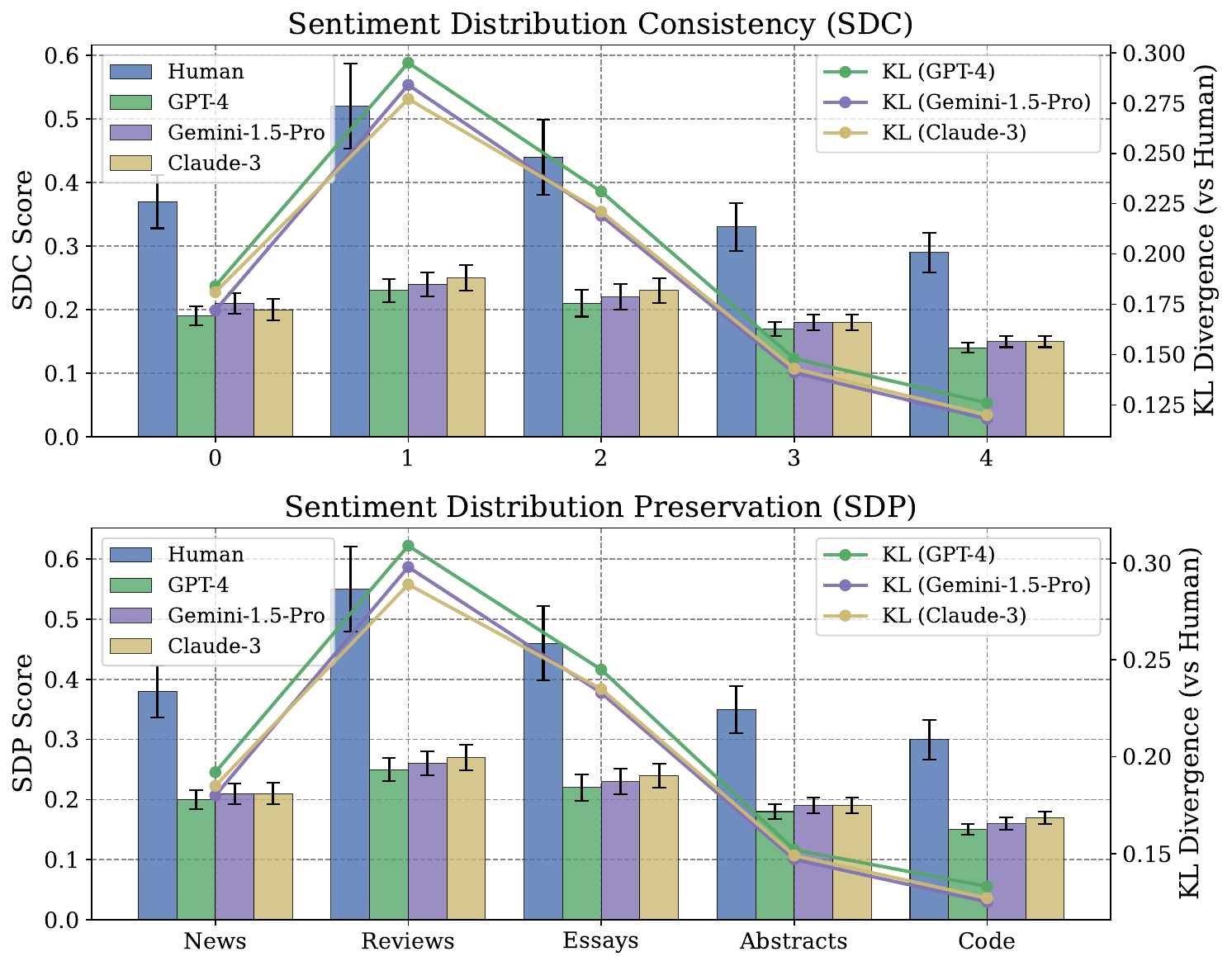}
    \caption{Cross-domain analysis of SDC and SDP scores and their KL divergence between LLM-generated and human-written text distribution.
    }
    \label{figure:SDC-SDP}
\end{figure}
To move beyond aggregate detection accuracy and better understand the behavioral basis of \ourmethod, we further examine how sentiment stability differs between human-written and LLM-generated texts across five representative domains.
As shown in~\autoref{figure:SDC-SDP}, a consistent asymmetry emerges in both SDC and SDP: human-written texts exhibit substantially greater variability, whereas LLM-generated texts, including those produced by \GPTFourAll, \GeminiOneFivePro, and \Claude, remain much more concentrated under the same stylistic and structural perturbations.
This contrast appears not only in mean values, but also in the overall shape of the empirical distributions.
Specifically, SDC and SDP are computed for each sample according to~\autoref{methodology} and then aggregated over large sets of human-written and LLM-generated texts.
The resulting distributions show that human writing is associated with broader dispersion, while machine-generated writing tends to cluster more tightly, indicating stronger rigidity under controlled rewriting.

To further quantify this difference, we estimate the KL divergence between the human and LLM-generated distributions, which provides a global measure of the expressive gap between the two sources.
The observed divergence supports a consistent interpretation: compared with human authors, LLMs preserve a more centralized affective profile even after rewriting, with reduced expressive fluctuation under controlled transformations.
More importantly, this tendency remains visible across all evaluated text types rather than being confined to a single genre.
This consistency strengthens the interpretation of sentiment stability as a behavior-level signal rather than a superficial artifact tied to particular datasets.
In this sense, SDC and SDP not only improve classification performance, but also provide an interpretable account of how LLM-generated texts differ from human-written texts under stylistic perturbation.

\textbf{\textit{(III) Emotional shift rate reveals low-variance expression patterns of LLMs.}}
\begin{figure}[!t]
    \centering
    \includegraphics[width=\linewidth]{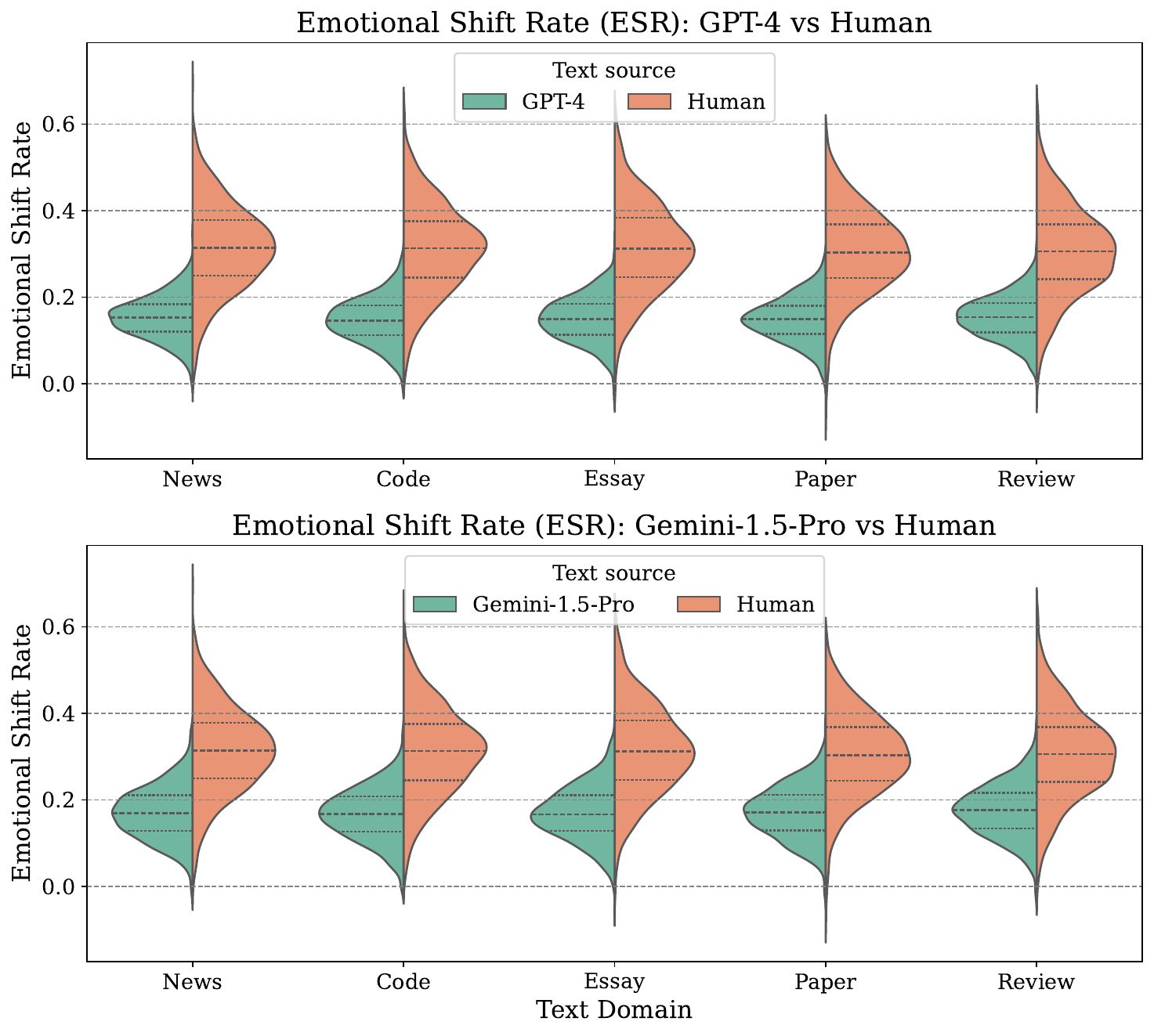}
    \caption{Illustration of comparing emotional shift rate distributions between LLM-generated texts (\GPTFourAll, \GeminiOneFivePro) and human-written texts across five domains.}
    \label{fig:esr_violin}
\end{figure}
Beyond lexical or statistical cues, we further investigate a dynamic behavioral signature, the Emotional Shift Rate (ESR). 
ESR quantifies emotional fluctuation between adjacent sentences in a text. 
It is motivated by the hypothesis that LLM-generated writing tends to maintain a more consistent affective tone, whereas human writing exhibits higher affective dynamics. 
To compute ESR, we segment each text sample into sentences. 
We then assign a sentiment polarity score $s_i$ to each sentence by prompting a fixed LLM-based scorer with the same sentiment evaluation instruction across all domains and sources. ESR is defined as the mean absolute difference between adjacent sentiment scores:
\begin{equation}
    \mathrm{ESR}(x) = \frac{1}{n - 1} \sum_{i=1}^{n - 1} |s_{i+1} - s_i|
\end{equation}
where $n$ is the number of sentences in the text $x$. We compute ESR distributions for 100 human-written and 100 LLM-generated texts per domain across five domains.

As shown in~\autoref{fig:esr_violin}, LLM-generated texts consistently exhibit lower ESR values, with a mean of approximately 0.15, while human-written counterparts have a mean ESR of approximately 0.31. 
This pronounced difference indicates that LLM outputs maintain a more consistent affective tone across sentences, suggesting inherent stability in their sentence-level sentiment patterns. 
This gap remains visible even in relatively neutral domains such as news articles and academic abstracts.
In these settings, although the affective space is constrained, human authors still introduce measurable dynamic variation due to changes in narrative focus and stylistic choices. 
By contrast, LLM outputs tend to show a more stable sentence-level affective profile under the same scoring procedure. 
Overall, ESR provides an interpretable signal that captures sentence-level sentiment stability, rather than relying on surface cues.

\begin{figure*}[!t]%
    \centering
    \includegraphics[width=\linewidth]{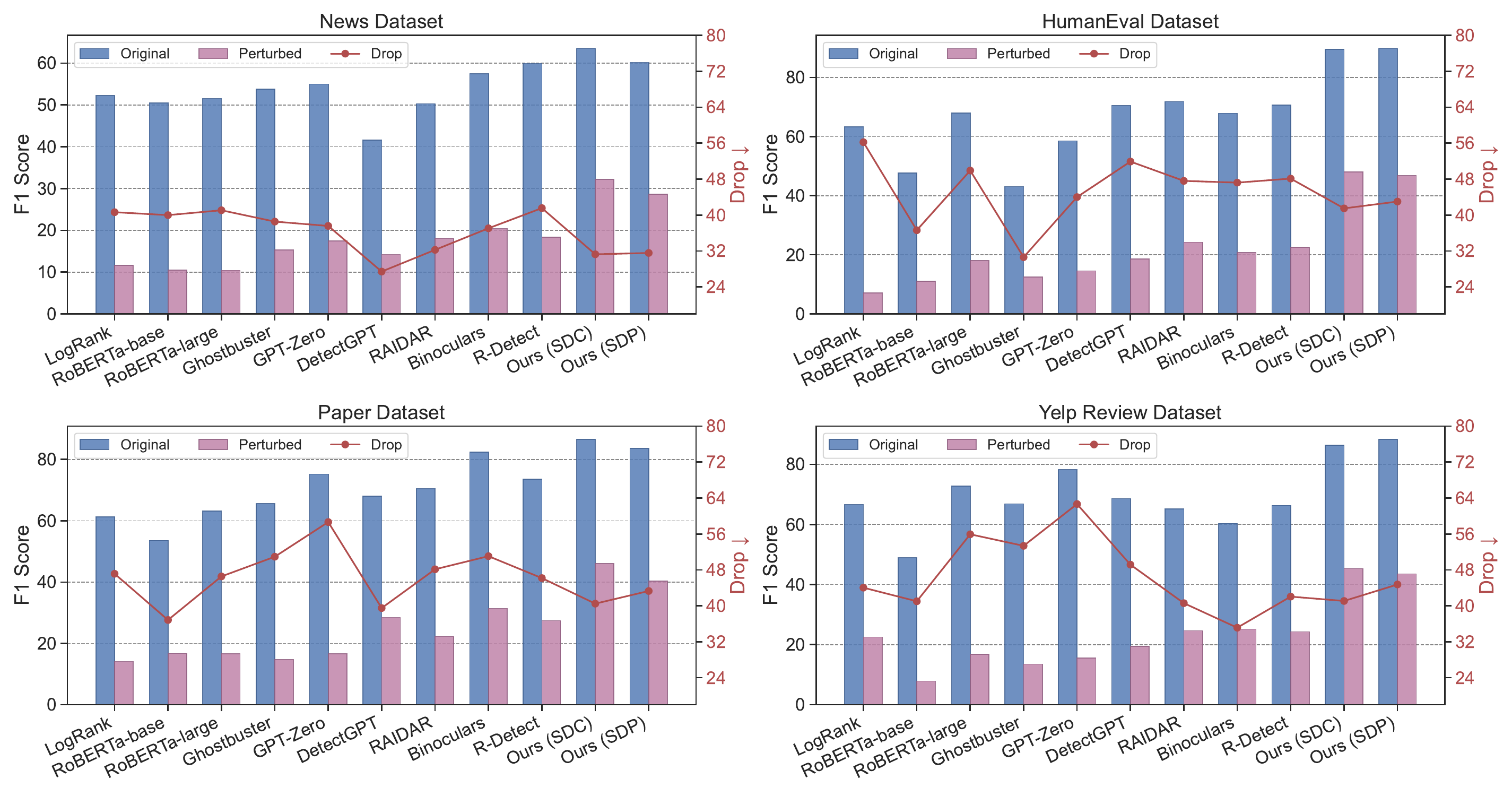}
    \caption{ Detecting F1 score degradation due to adversarial perturbation.
    The first and second bars for each method show the original and post-perturbation scores, respectively.
    The line chart highlights the F1 score drop caused by the perturbation.}
    \label{figure:perturbation}
\end{figure*}

% ----------------
\begin{table}[!t]
    \centering
    \small
    \caption{Examples of sentiment vectors generated via $F_2$.}
    \label{table:appendix_f2_examples}
    % \resizebox{\linewidth}{!}{
    \begin{tabular}{l|ccc}
        \toprule
        \textbf{Rewritten Text} & \textbf{Neg} & \textbf{Neu} & \textbf{Pos} \\
        \midrule
        The food was somewhat bland & 0.13 & 0.76 & 0.11 \\
        It felt a bit underwhelming & 0.18 & 0.70 & 0.12 \\
        Service was terribly slow & 0.72 & 0.22 & 0.06 \\
        Absolutely loved the dessert & 0.05 & 0.20 & 0.75 \\
        \bottomrule
    \end{tabular}
    % }
\end{table}
% ----------------

\subsubsection*{\textbf{(IV) Detailed Analysis of $F_2$}}
To support our Sentiment Distribution Preservation (SDP) strategy in \ourmethod, we design a mapping function $F_2$ that captures the emotional profile of a text based on multiple structure-preserving rewrites. 
Given an input sample, we apply $N$ sentiment-invariant rewriting prompts (e.g., paraphrasing, person shifting, length variation) to produce $N$ semantically equivalent variants.
Each rewritten sample is processed by a 3-class sentiment classifier, yielding a probability vector $[p_{\text{neg}}, p_{\text{neu}}, p_{\text{pos}}]$, as illustrated in Table~\ref{table:appendix_f2_examples}.
By concatenating these vectors, $F_2$ forms a $3 \times N$ sentiment distribution matrix, which we use to derive detection features such as KL divergence.
These representations form the basis for divergence-based signals, which reflect the degree of sentiment fluctuation across rewrites, a key indicator distinguishing human-written and LLM-generated text.

\vspace{-0.2\baselineskip}
\begin{tcolorbox}[colback=gray!10, colframe=gray!90, boxrule=0.4pt, arc=1mm, left=4pt, right=4pt, top=4pt, bottom=4pt]
\textbf{Finding 2: Cross-Domain Stability and Interpretability}\\
(1) The sentiment-invariant signal is insensitive to domain-specific writing, enabling stable detection across domains.\\
(2) Sentiment-invariant patterns are interpretable signals for detection across distributional and sequential analyses.
\end{tcolorbox}
\vspace{-0.2\baselineskip}

\subsection{Robustness Analysis and Ablation Study} 
\textbf{\textit{(I) Robustness Against Adversarial Perturbation.}}

We use TextAttack\footnote{https://github.com/QData/TextAttack/} to implement adversarial perturbation attacks that aim to evade detection through lexical-level modifications.
In this setting, only LLM-generated texts are perturbed, and the effect of the attack is evaluated by comparing the original F1 score, the post-perturbation F1 score, and the corresponding performance drop.
As shown in~\autoref{figure:perturbation}, adversarial perturbation reduces detection performance for all methods, confirming that word-level attacks remain a practical threat to text provenance detection.
However, the degradation is not uniform across detectors.
Methods that rely more heavily on token-level statistics, local likelihood irregularities, or surface cues tend to suffer larger drops once lexical substitutions alter the original realization of the input.
This pattern is particularly visible for GPT-Zero, DetectGPT, Binoculars, and RAIDAR, whose performance margins shrink substantially.

By contrast, \ourmethod shows a more controlled degradation pattern across the evaluated datasets.
Although its post-attack performance also decreases, the reduction is consistently less severe than that of the strongest baselines, and the method preserves a clearer margin in structurally constrained domains such as \datasetcode and \datasetpaper.
This behavior is consistent with the design of \ourmethod, whose decision signal is derived from distributional changes in sentiment features under controlled rewriting rather than from fragile lexical artifacts alone.
As a result, local substitutions may distort surface form, but do not remove the higher-level feature captured by the SDC/SDP measures, allowing \ourmethod to retain stronger detection capability under adversarial perturbations.

\textbf{\textit{(II) Robustness Against Cross-lingual Evasion.}} 
In practical security forensics, adversaries often employ round-trip translation (English $\rightarrow$ Chinese $\rightarrow$ English) as a sophisticated obfuscation technique to erase local statistical signatures and specific generative content of LLMs. 
Furthermore, existing studies suggest that machine translation can act as a semantic smoothing filter, potentially perturbing the emotional expression and stylistic nuance of the original text \cite{troiano2020lost}. 
To verify the stability of \ourmethod in such cross-lingual evasion scenarios, we conduct a robustness evaluation using machine translation as a form of semantic-preserving perturbation.

As shown in~\autoref{tab:cross_lingual}, we summarize the detection scores and performance drop under round-trip translation (EN-ZH-EN) on both the \datasetnews and \datasetreview datasets.
We observe that although the translation cycle introduces lexical variations and syntactic restructuring, \ourmethod maintains high detection stability.
While statistical detectors typically suffer significant performance degradation due to sensitivity to translation-induced noise, \ourmethod captures emotional response inertia, a high-level behavioral signal that remains invariant under linguistic transformations.
As shown in the table, the F1 score drop for \ourmethod is limited to only about 7.0\%, consistently lower than all baseline methods, demonstrating strong forensic reliability even when LLM-generated content undergoes cross-lingual rewriting.

\begin{table}[!t]
    \centering
    \caption{ Robustness of detection under Machine Translation (Back-translation). 
    We compare the original (EN) performance with the cross-lingual (EN-ZH-EN) performance and report the Drop$\downarrow$.}
    \label{tab:cross_lingual}
    \setlength{\tabcolsep}{3pt}
    \resizebox{\columnwidth}{!}{
    \begin{tabular}{lcccccc}
        \toprule
        \multirow{2}{*}{\textbf{Method}} & \multicolumn{3}{c}{\textbf{\datasetnews Dataset}} & \multicolumn{3}{c}{\textbf{\datasetreview Dataset}} \\ \cmidrule(lr){2-4} \cmidrule(lr){5-7}
        & \textbf{Original} & \textbf{Translated} & \textbf{Drop$\downarrow$} & \textbf{Original} & \textbf{Translated} & \textbf{Drop$\downarrow$} \\ \midrule
        % Rank & 0.821 & 0.710 & 13.5\% & 0.842 & 0.720 & 14.5\% \\
        LogRank & 0.833 & 0.719 & 13.7\% & 0.851 & 0.724 & 14.9\% \\
        \rowcolor{gray!10} RoBERTa-base & 0.841 & 0.758 & 9.9\% & 0.852 & 0.763 & 10.4\% \\
        RoBERTa-large & 0.862 & 0.793 & 8.0\% & 0.872 & 0.792 & 9.2\% \\
        \rowcolor{gray!10} Ghostbuster & 0.854 & 0.782 & 8.4\% & 0.863 & 0.781 & 9.5\% \\
        GPT-Zero & 0.812 & 0.706 & 13.1\% & 0.832 & 0.715 & 14.1\% \\
        \rowcolor{gray!10} DetectGPT & 0.843 & 0.752 & 10.8\% & 0.853 & 0.753 & 11.7\% \\
        Fast-DetectGPT & 0.871 & 0.781 & 10.3\% & 0.881 & 0.783 & 11.1\% \\
        \rowcolor{gray!10} Binoculars & 0.883 & 0.799 & 9.5\% & 0.891 & 0.798 & 10.4\% \\
        RAIDAR & 0.861 & 0.798 & 7.3\% & 0.871 & 0.797 & 8.5\% \\ 
        \rowcolor{gray!10} R-Detect & 0.872 & 0.808 & 7.3\% & 0.882 & 0.809 & 8.3\% \\
        \hline
        \textbf{DSIPA (SDC)} & 0.885 & 0.824 & 6.9\% & 0.893 & 0.828 & 7.3\% \\
        \textbf{DSIPA (SDP)} & 0.892 & 0.835 & 6.4\% & 0.898 & 0.838 & 6.7\% \\ 
        \bottomrule
    \end{tabular}
    }
\end{table}

\textbf{\textit{(III) Effect of Input Text Length.}}
\begin{figure}[!t]
    \centering
    \includegraphics[width=\linewidth]{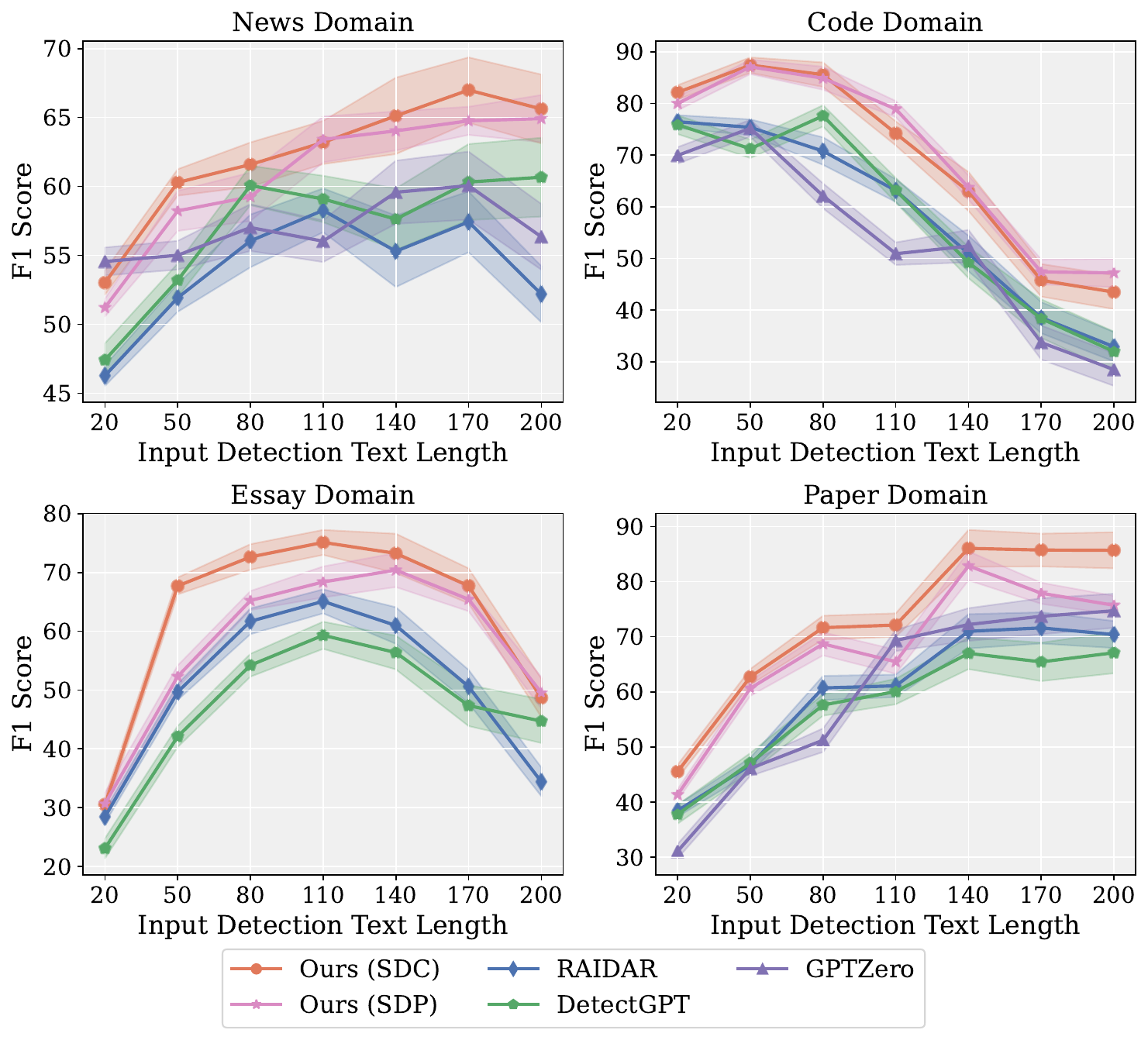}
    \caption{Varying length detection comparison results of \ourmethod and baselines on the \datasetnews, \datasetcode, \datasetessay, and \datasetpaper datasets.}
    \label{figure:length}
\end{figure}

Experiments conducted on \GPTThreeFiveTurbo reveal that detection performance varies with input text length. 
Overall, longer inputs tend to yield higher F1 scores, especially in the \datasetpaper and \datasetnews domains, where the performance steadily improves as length increases. 
The richer content in longer texts allows \ourmethod to capture more consistent sentiment-invariant patterns and extract stronger behavioral signals, resulting in more reliable provenance attribution.
In contrast, the \datasetcode and \datasetessay domains exhibit a non-monotonic trend, where F1 scores increase initially but decline beyond a certain length. 
This suggests that overly long texts may dilute the sentiment-invariant signal, as the accumulation of redundant structural content reduces the per-sample divergence captured under low-emotional rewriting.
At the other end of the length spectrum, short inputs pose a greater challenge due to limited sentiment fluctuation and sparse stylistic cues, providing less behavioral information for robust detection. 

As LLMs improve, their short outputs increasingly mimic human-written text, making origin attribution more difficult.
For instance, as reflected in the ESR analysis in \autoref{fig:esr_violin}, although samples from different sources share remarkably similar surface tone, they differ in deep emotional dynamics, which \ourmethod exploits to maintain detection reliability.
As shown in~\autoref{figure:length}, our method consistently outperforms existing baselines across all domains and length settings, even at a 20-token limit. 
This demonstrates that \ourmethod effectively leverages cross-segment sentiment-invariant features effectively to extract high-level behavioral signals, maintaining strong performance even in extremely short texts, while baselines relying primarily on surface cues or local token statistics experience greater performance degradation.

To further evaluate the scalability of \ourmethod in the context of modern LLMs that support extremely long document-level generation, we extend the evaluation to longer input segments of 512 and 1024 tokens on \GPTFiveTwoAll, which offers stronger long-context generation capability than the GPT-3.5-turbo used in the preceding length analysis.
As summarized in~\autoref{tab:long_context}, the detection accuracy of \ourmethod continues to scale positively with input length across all three domains, demonstrating a consistent and stable upward trend well beyond the shorter observation windows.
This behavior aligns with the emotional response inertia captured by \ourmethod, as longer inputs provide richer and more continuous contextual cues, thereby reducing the interference of random semantic fluctuations that are common in short texts.
The gain is most pronounced in the academic paper domain, where the inherent logical coherence and stylistic consistency of scholarly writing are more fully captured as the observation window expands.
These results confirm that \ourmethod remains robust and effective on long-form content, which is crucial for practical large-scale document detection scenarios.

\begin{table}[!t]
    \centering
    \caption{Detection performance of \ourmethod with extended input lengths on \GPTFiveTwoAll, across the \datasetnews, \datasetessay, and \datasetpaper domains.}
    \label{tab:long_context}
    \resizebox{\columnwidth}{!}{
    \begin{tabular}{lccccc}
        \toprule
        \textbf{Task} & \textbf{64 Token} & \textbf{128 Token} & \textbf{256 Token} & \textbf{512 Token} & \textbf{1024 Token} \\ \midrule
        News & 0.785 & 0.842 & 0.889 & 0.908 & 0.921 \\
        Paper & 0.753 & 0.821 & 0.893 & 0.916 & 0.931 \\
        Essay & 0.768 & 0.830 & 0.882 & 0.903 & 0.917 \\ \bottomrule
    \end{tabular}
    }
\end{table}

\textbf{\textit{(IV) Computational Efficiency Comparison.}}
While \ourmethod introduces a rewriting mechanism to capture emotional response inertia, it is crucial to verify that this process does not incur prohibitive computational costs.
To evaluate the practical deployability of our method, we conduct a runtime analysis comparing \ourmethod with representative baselines. 
The experiment is performed on a standard validation set with a fixed input length of approximately 512 tokens per sample. 
For \ourmethod, the number of rewrite iterations is set to the default $N=5$, which balances detection performance and runtime. 
All methods are evaluated in the same hardware environment to ensure a fair comparison.
\begin{table}[!t]
    \centering
    \caption{Computational efficiency analysis. Average runtime (Avg. Runtime, in seconds) per 512-token sample with $N=5$, measured under an identical hardware environment for all methods.}
    \label{tab:efficiency_time}
    \footnotesize
    \setlength{\tabcolsep}{5pt}~
    \renewcommand{\arraystretch}{1.15}
    % \resizebox{\columnwidth}{!}{
    \begin{tabular}{llc}
        \toprule
        \textbf{Method} & \textbf{Category} & \textbf{Runtime(s)} \\
        \midrule
        Rank            & Probability-based/Ranking & 0.02 \\
        LogRank         & Probability-based/Ranking & 0.04 \\
        RoBERTa-large   & Classifier-based  & 0.38 \\
        DetectGPT       & Perturbation-based/Adversarial & 18.62 \\
        Fast-DetectGPT  & Perturbation-based/Adversarial & 9.45 \\
        Binoculars      & Perturbation-based/Adversarial & 4.22 \\
        RAIDAR          & Rewrite-based & 3.76 \\
        \ourmethod & Behavioral/Sentiment Stability Pattern & \textbf{2.18} \\
        \bottomrule
    \end{tabular}
    % }
\end{table}

\begin{table}[!t]
    \centering
    \caption{Ablation study on the number of rewrite iterations ($N$) on \GPTFiveTwoAll, reporting token consumption, runtime, and detection F1 alongside cumulative token-and-runtime cost and incremental performance gains.}
    \label{tab:ablation_rewrite}
    \resizebox{0.82\columnwidth}{!}{
    \begin{tabular}{cccccc}
        \toprule
        \textbf{$N$} & \textbf{Tokens} & \textbf{Runtime (s)} & \textbf{F1} & \textbf{$\Delta$Cost} & \textbf{$\Delta$F1} \\ \midrule
        1 & ~512 & 0.87 & 86.3 & - & - \\
        3 & 1536 & 1.92 & 91.7 & +200\% & +5.4 \\
        5 & \textbf{2560} & \textbf{2.98} & \textbf{93.2} & \textbf{+400\%} & \textbf{+1.5} \\
        7 & 3584 & 3.95 & 93.5 & +600\% & +0.3 \\
        9 & 4608 & 4.91 & 93.6 & +800\% & +0.1 \\ \bottomrule
    \end{tabular}
    }
\end{table}
As reported in \autoref{tab:efficiency_time}, \ourmethod requires an average of 2.18 seconds per 512-token sample, positioning it among the more efficient methods that involve model inference.
Compared with perturbation-based and inference-heavy detectors, \ourmethod achieves speedups of nearly $9\times$ over DetectGPT and over $4\times$ over Fast-DetectGPT while maintaining superior detection accuracy. 
Although \ourmethod incurs higher runtime than lightweight statistical baselines such as Rank and LogRank owing to the additional inference overhead introduced by the rewriting mechanism, these baselines fall considerably short of \ourmethod in detection accuracy across all evaluated domains, yielding a far less favorable trade-off between accuracy and efficiency in practice.
These results demonstrate that \ourmethod achieves a favorable balance between detection accuracy and computational overhead, supporting its deployment in forensic investigation and large-scale content attribution scenarios where both reliability and efficiency are required.

\textbf{\textit{(V) Ablation Study on Rewrite Iterations.}} 
The number of rewrite iterations ($N$) is a key hyperparameter in \ourmethod, affecting both the stability of sentiment features after rewriting and the overall computational cost (jointly measured by token consumption and runtime). 
To characterize the trade-off between detection performance and resource consumption, we vary $N$ from 1 to 9 on the \datasetnews dataset, with inputs truncated to 200 tokens following our main protocol. 
For each setting, we report the average per-sample token consumption (including input and output tokens from rewriting), the average runtime, and the detection F1 score. 

As shown in \autoref{tab:ablation_rewrite}, token consumption and runtime grow linearly with $N$, as each iteration requires an independent \GPTFiveTwoAll generation call. 
Detection performance exhibits a clear diminishing-returns pattern: F1 improves substantially at small $N$, stabilizes around $N=5$, and yields only marginal gains beyond this point. 
This is because five iterations suffice to form a stable aggregation of sentiment features, capturing sentiment-invariant differences between LLM-generated and human-written text. 
Excessive iterations increase cost without meaningful gains, while insufficient iterations yield less reliable estimates due to limited aggregation stability. 
Thus, we adopt $N=5$ as the default to balance cost and accuracy.

\vspace{-0.2\baselineskip}
\begin{tcolorbox}[colback=gray!10, colframe=gray!90, boxrule=0.4pt, arc=1mm, left=4pt, right=4pt, top=4pt, bottom=4pt]
\textbf{Finding 3: Robustness and Efficiency.}\\
(1) \ourmethod sustains stable under adversarial perturbation, cross-lingual rewriting, and input length variation.\\
(2) \ourmethod reaches its performance ceiling with limited computational overhead, without requiring further scaling.
\end{tcolorbox}
\vspace{-0.2\baselineskip}

\section{Discussion and Conclusion}
This work presents \ourmethod, a training-free, black-box framework for detecting LLM-generated text by measuring sentiment-invariant divergence under controlled low-emotional rewriting. Through two complementary metrics, \textit{Sentiment Distribution Consistency} and \textit{Sentiment Distribution Preservation}, \ourmethod operationalizes what we term \textit{emotional response inertia} as an interpretable behavioral signal, without requiring token probabilities, watermarking, or white-box access.
Across five domains and multiple LLM families, the results show strong detection accuracy and stable cross-domain behavior, with robustness under paraphrasing, adversarial perturbations, and input-length variation. These findings support the use of affective stability under rewriting as a practical cue for provenance attribution.
Despite these strengths, the method can be less reliable for very short inputs and for domains with limited stylistic flexibility, where sentiment variation is intrinsically constrained. Moreover, as future LLMs are optimized to better emulate human affective dynamics, the discriminative margin induced by sentiment stability may narrow. Future work will study hybrid behavioral signals that integrate sentiment stability with complementary cues, such as syntactic variability and revision dynamics, to improve robustness under stronger adaptive attacks.

\bibliographystyle{IEEEtran}
\bibliography{ref}

\end{document}